\documentclass{article}
\usepackage{arxiv}

\usepackage[utf8]{inputenc} 
\usepackage[T1]{fontenc}    
\usepackage{cite}
\usepackage{amsmath,amssymb,amsfonts,amsthm}
\usepackage{algorithmic}
\usepackage{graphicx}
\usepackage{textcomp}
\usepackage{xcolor}
\usepackage{url}
\usepackage{booktabs}  
\usepackage{pifont}
\usepackage{hyperref}
\usepackage{tablefootnote}
\usepackage{comment}
\usepackage{subcaption}
\usepackage{longtable}
\usepackage{textcase}
\usepackage{float}
\usepackage{orcidlink}
\usepackage{makecell}
\usepackage{appendix}
\graphicspath{ {./images/} }

\newcommand{\cfclust}{\sc{CfClust}}
\newcommand{\dice}{\sc{dice}}
\newcommand{\prt}{\sc{prt}}
\newcommand{\Iris}{{Iris}}
\newcommand{\Wine}{{Wine}}
\newcommand{\two}{{2D}}
\newcommand{\three}{{3D}}
\newcommand{\pendigits}{{Pendigits}}

\title{Counterfactual Explanations for $k$-means and Gaussian Clustering}

\author{
Georgios Vardakas \\
Department of Computer Science \& Engineering\\
University of Ioannina\\
Ioannina, Greece \\
\texttt{g.vardakas@uoi.gr} \\
\And
Antonia Karra\\
Department of Computer Science \& Engineering\\
University of Ioannina\\
Ioannina, Greece \\
\texttt{a.karra@uoi.gr}\\
\And
Evaggelia Pitoura\\
Department of Computer Science \& Engineering\\
University of Ioannina\\
Ioannina, Greece \\
\texttt{pitoura@cs.uoi.gr} \\
\And
Aristidis Likas\\
Department of Computer Science \& Engineering\\
University of Ioannina\\
Ioannina, Greece \\
\texttt{arly@cs.uoi.gr} \\
}

\begin{document}
\maketitle
\begin{abstract}
Counterfactuals have been recognized as an effective approach to explain classifier decisions. Nevertheless, they have not yet been considered in the context of clustering. In this work, we propose the use of counterfactuals to explain clustering solutions. First, we present a general definition for counterfactuals for model-based clustering that includes plausibility and feasibility constraints. Then we consider the counterfactual generation problem for $k$-means and Gaussian clustering assuming Euclidean distance. Our approach takes as input the factual, the target cluster, a binary mask indicating actionable or immutable features and a plausibility factor specifying how far from the cluster boundary the counterfactual should be placed. In the $k$-means clustering case, analytical mathematical formulas are presented for computing the optimal solution, while in the Gaussian clustering case (assuming full, diagonal, or spherical covariances) our method requires the numerical solution of a nonlinear equation with a single parameter only. We demonstrate the advantages of our approach through illustrative examples and quantitative experimental comparisons.
\end{abstract}


\section{Introduction}
Explainable AI is necessary for developing trustworthy and transparent machine learning models, enabling understanding and effectively using these systems. Explainability methods can be distinguished into global and local explanation approaches~\cite{ribeiro2016should}. Global explanation methods aim to provide insights into the overall behavior of a model, often using techniques such as decision trees or rule-based systems. Such methods summarize the model’s behavior across the entire dataset. In contrast, local explanation methods focus on individual predictions or decisions, offering insights into specific outcomes. Examples of local approaches include counterfactual explanations~\cite{wachter2017counterfactual}, which identify the minimal changes needed to alter a model's decision, and saliency maps~\cite{adebayo2018sanity}, which highlight the most influential features of a particular prediction.

Counterfactual explanations (CFEs) have been established as an effective \emph{local explanation} approach for explaining decisions of machine learning models~\cite{guidotti2024counterfactual,verma2024counterfactual,artelt2019computation}. The highly referenced Alice's example for loan qualification offers an intuitive insight into the counterfactual idea: Suppose that there are two possible model outcomes: $c = \texttt{`negative'}$, the applicant does not qualify for the loan, and $c' = \texttt{`positive'}$, the applicant does qualify for the loan. Alice did not receive the loan she applied for, since $f(y) = c$, meaning that the input vector $y$ corresponding to Alice did not produce the desired output $c'$. An explanation for such a decision is needed to help Alice get the loan in the future: what is the minimum required change that Alice should make (in terms of income, education, etc.) in order to qualify for the loan. 

More generally, this is the kind of explanation counterfactuals give to a classification decision: \emph{What is the minimum change that can be applied to $y$ (producing $z$) so that the model output changes from $f(y) = c$ to $f(z) = c'$?} The definition of counterfactual in the context of classification is the following:

Given a classifier $f$ that outputs the decision $c = f(y)$ for a factual instance $y$, a counterfactual explanation consists of an instance $z$ such that i) the decision for $f$ on $z$ is different from $c$, ie., $f(z) \neq c$, and ii) the difference between $y$ and $z$ is minimum~\cite{guidotti2022counterfactual,verma2024counterfactual}.

Moreover, several interesting properties of counterfactuals for classification have been defined, such as \emph{actionability} (some features are not allowed to change) and \emph{plausibility} (the counterfactual should lie inside the data manifold).

Although considerable research work has been devoted to counterfactuals to explain classification decisions, to the best of our knowledge, no research results are available for computing counterfactuals to explain \emph{clustering} solutions. In the following sections we attempt to fill this gap by presenting novel definitions and methods. More specifically:
\begin{itemize}
	\item We present a general definition of counterfactuals for clustering assuming that each cluster is modeled using a probability density.
	\item $k$-means clustering solutions can be considered as a special case of the above framework.
	\item We introduce easy to compute (non-iterative) methods for generating counterfactuals considering \emph{Euclidean distance} between factual and counterfactual for the cases of:
	\begin{itemize}
		\item clusters obtained by $k$-means clustering
        \item Gaussian clusters (with full, diagonal and spherical covariances).
	\end{itemize}
    \item Our approach also accounts for:
    \begin{itemize}
    	\item actionable and immutable features
    	\item counterfactual plausibility (moving from to the cluster boundary towards regions of higher cluster density).
    \end{itemize}
\end{itemize}
To assess the effectiveness of the proposed techniques, we have considered synthetic and real datasets. We compare the generated counterfactuals to those provided by considering an indirect approach that treats the problem as classification and exploits the tools and methods available for explaining classifier decisions.

The rest of the paper is organized as follows. In Section~\ref{sec:related} we briefly discuss related work on  explainable clustering. In Section~\ref{sec:definition} we present a counterfactual definition for model-based clustering followed by the general framework for counterfactual computation that is proposed in Section~\ref{sec:framework}. In Section~\ref{sec:k-means} the analytical formulas for computing counterfactuals in the $k$-means case are derived, while in Section~\ref{sec:gaussian} we derive the equations to be solved in the case of Gaussian clusters with full, diagonal and spherical covariance matrices. In Section~\ref{sec:experiments} we report the details and results of our comparative experimental study, while Section~\ref{sec:conclusions} provides conclusions and future research directions. 

\section{Related Work}
\label{sec:related}

Explainable methods for clustering focus on \emph{global explanations}. In particular, they build \emph{decision tree}  models that directly provide interpretations in the form of decision rules. Several methods have been proposed to build decision trees to explain clustering by utilizing indirect or direct approaches. 

The \emph{indirect} global explanation methods,  typically follow a two-step procedure: first, they obtain cluster labels using a clustering algorithm, such as $k$-means,
and then they apply a supervised decision tree algorithm to build a decision tree that interprets the resulting clusters. For example, in~\cite{laber2023shallow} labels obtained from $k$-means are used as a preliminary step in tree construction. Similarly in~\cite{tavallali2021k} the centroids derived from $k$-means are also involved in splitting procedures. 

\emph{Direct} global explanation methods integrate decision tree construction and partitioning into clusters. Many of them follow the typical top-down splitting procedure used in the supervised case
but exploit unsupervised splitting criteria, e.g., compactness of the resulting subsets~\cite{bertsimas2021interpretable} or unimodality criteria~\cite{chasani2024unsupervised}.

To the best of our knowledge, the \emph{local explanation} problem has not yet been addressed in the case of clustering.
In this paper, we introduce counterfactual-based local explanations for clustering.
Counterfactual explanations are local explanations proposed for classification \cite{wachter2017counterfactual, verma2024counterfactual,guidotti2024counterfactual}.
We present a novel definition for clustering and novel counterfactual generation approaches for clustering.
Local explanations enable a deeper understanding of individual assignments and their relationship to cluster characteristics.
Furthermore, in addition to explaining the cluster assignment of  a specific instance, counterfactual explanations also provide 
the minimum feature changes for the instance to be assigned to a different cluster.



\section{Counterfactual definition for model-based clustering}
\label{sec:definition}
We assume the general clustering problem with $C_1, ..., C_M$ clusters. We are given the probability density $p_k(x)$ for each cluster $C_k$, $k=1, ..., M$. Let also $(\pi_1,...,\pi_M)$ be the prior probability vector of the clusters. Typically $\pi_k$ is set equal to the cluster frequencies ($\pi_k=N_k/N$) or equal to $1/M$. Based on the \emph{cluster assignment rule}, an example $x$ is assigned to cluster  $C_\ell$ for which $\pi_\ell p_\ell(x)$ is maximum \cite{kaufman2009finding}.

 \begin{figure}[ht]
    \centering
    \includegraphics[width=0.65\linewidth]{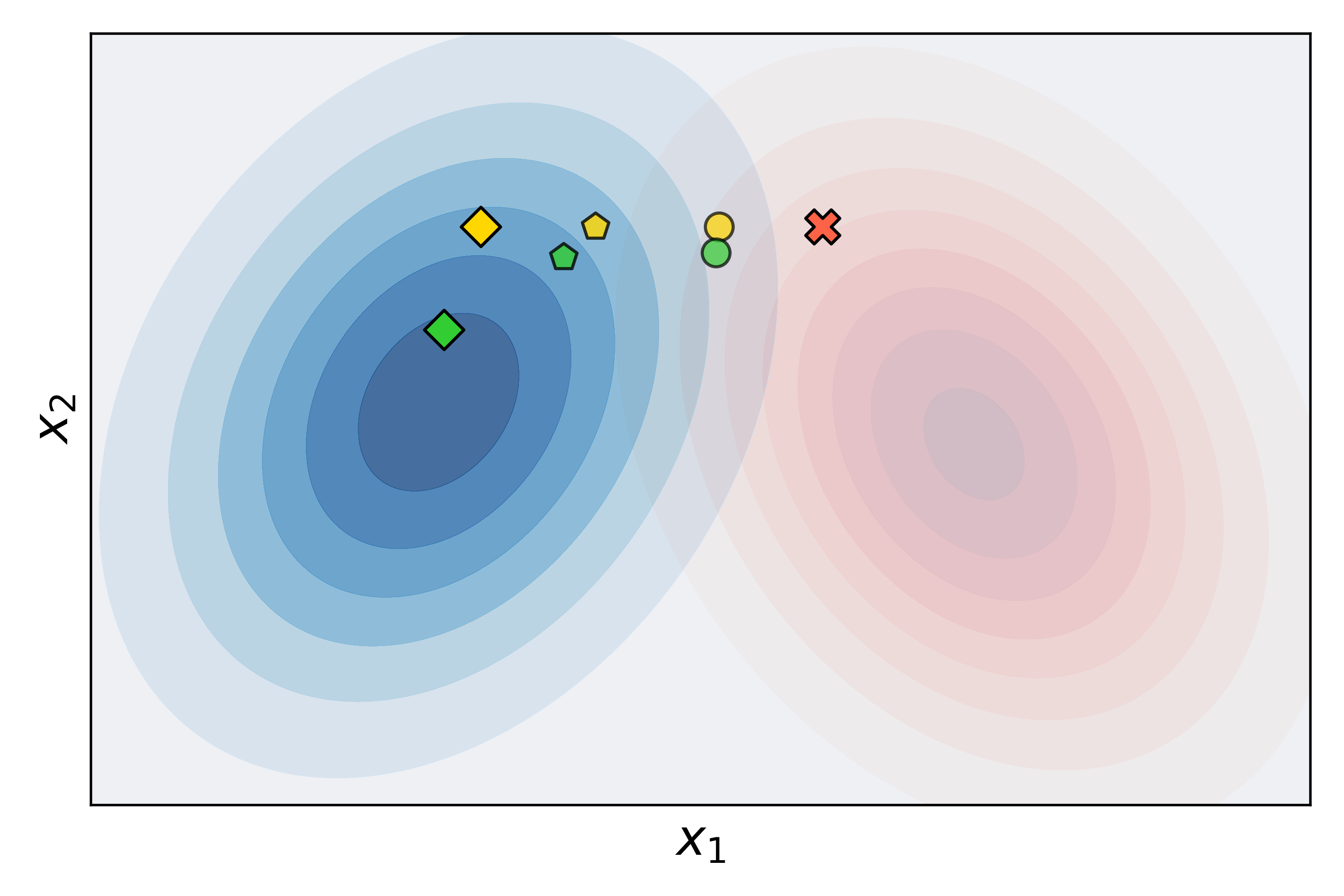}
    \caption{Illustration of the counterfactuals generated (green and yellow points) for the given factual y (marked as red cross) under the assumption of Euclidean distance and different feasibility and plausibility constraints.}
    \label{fig:definition}
\end{figure}

In order to compute a counterfactual explanation (CFE) $z$ for a given example (factual) $y$, we need to specify a \emph{preference density} $r(x|y)$ which should be a \emph{unimodal distribution} with mode at $y$ expressing the \emph{preference} for $x$ to become a counterfactual of $y$. This preference density definition is very general and flexible and also implements feature actionability, as will be discussed later. 
\\ 
\noindent \textbf{CFE Definition}: Given a factual $y$ of cluster $C_k$, its \emph{counterfactual explanation} (CFE) $z$ is defined as the solution to the following constrained optimization problem:
\begin{equation}
\label{eq.def-r}
	z = \arg\max_x r(x|y) \text{ ($z$ has maximum preference given $y$)}
\end{equation}
\begin{center}
\noindent subject to the constraint:
\end{center}
\begin{equation}
	C_\ell \neq C_k \text{ where } C_\ell = \arg\max_{C_m} \pi_m p_m (z) 
\end{equation}
In the above formulation, $C_\ell$ is the cluster label of $z$, taking into account the cluster assignment rule. The preference density $r(x|y)$ should be a unimodal function (with peak at the factual $y$) that should decrease as the distance $d(x,y)$ between $x$ and $y$ increases. The exponential form ($r(x|y)=exp(-d(x,y))$) is a viable option, although other functional forms could be devised. Note that, if $r(x|y)$ is monotonically decreasing with respect to $d(x,y)$, then the maximization of $r(x|y)$ is equivalent to the minimization of $d(x,y)$.

Extending the previous definition, it is possible to extend the set of constraints. More specifically a \emph{plausibility constraint} can be specified on the counterfactual $z$: 
\begin{equation}
p_\ell (z) > \delta \text{ (plausibility)} \\
\end{equation}
where the parameter $\delta>0$ is a threshold on the local cluster density of $z$. Adjusting $\delta$ allows the counterfactual to lie in regions of sufficient cluster density, i.e. it does not constitute an outlier. 

Moreover, by appropriately defining $r(x|y)$, we can introduce \emph{feasibility} (or actionability) constraints: 
\begin{equation}
r(z|y)>0 \text{ (feasibility)} 
\end{equation}
ie., impose constraints on the allowed feature values of counterfactual $z$. 
If for example the values of the $i$-th feature are not allowed to change, we can set $r(x|y)=0$ for all $x$ with $x_i \neq y_i$. In this case, the feasibility constraint will prevent the generation of counterfactual $z$ with $z_i \neq y_i$.

In Fig.~\ref{fig:definition}, we see an example for the case of two clusters in two dimensions. For the given factual point $y$ (marked as red cross), we plot counterfactuals with a minimum Euclidean distance from $y$ under feasibility and plausibility constraints. Green points represent counterfactuals with both features actionable (allowed to change) obtained for increasing value of $\delta$. Yellow points represent counterfactuals with only the horizontal feature actionable obtained for increasing values of $\delta$.

In the following sections, we focus on \emph{Gaussian clusters}, i.e. the probability density of each cluster $i$ follows the Gaussian distribution $p_i(x)=N(x;\mu_i, S_i)$ 
where $\mu_i$ is the mean (cluster center) and $S_i$ is the covariance matrix of the distribution. It should be stressed that $k$-means clustering can be considered a special case of Gaussian clustering where all priors $\pi_i$ are equal and all covariance matrices are spherical with unit variance, thus $k$-means clustering is included in this framework.

A convenient property of a Gaussian cluster is that its density $p(x)$ is inversely proportional to the (Mahalanobis in the general case) 
distance of the cluster center. Therefore, considering two clusters $i$ and $j$, there exists a cluster boundary $B(i,j)$, i.e. a set of points $x$ for which $\pi_i p_i(x)= \pi_j p_j(x)$. Assuming that the cluster centers do not coincide, 
as we depart from the cluster boundary and move inside cluster $j$,  
the density $p_j$ increases, while the density $p_i$ decreases. Note also that cluster assignment is based on the maximum cluster density.
Consequently, given a factual point $y$ of cluster $i$ (ie.  $\pi_i p_i(y)> \pi_j p_j(y)$), it is ensured that its closest counterfactual $z$ 
(with respect to target cluster $j$) will be located on the cluster boundary, i.e., $z \in B(i,j)$. This fact constitutes the basis of our methods. Given a factual $y$, the counterfactual $z$ is computed
as its closest point that lies on the cluster boundary. 
This idea is then extended to address the case where the counterfactual is desired to be positioned not at the cluster boundary, 
but within a more dense region of the target cluster, in order to achieve higher plausibility. In this case, the user should specify a parameter that we call \emph{plausibility factor}.

\section{Counterfactual Computation: General Framework}
\label{sec:framework}
In all algorithms for counterfactual generation that are presented below, we consider \emph{pairs of clusters}: \emph{with $C_s$ denoting the source cluster of factual $y$ and $C_t$ the target cluster of counterfactual $z$.} The reason is that, if more than one counterfactual clusters exist (e.g. $C_2$, $C_3$ etc), we can apply the algorithms separately for each cluster pair $(C_s, C_t)$ $(t\neq s)$, compute the corresponding counterfactuals and keep the one with maximum preference (e.g. minimum distance from the factual).


The clustering information that should be given is the clustering model, ie. the two cluster centers in the $k$-means case or the parameters of the two Gaussians and the priors $\pi$ in the Gaussian clustering case. \emph{No data availability is required}. 

We assume a preference function of the form:
\begin{equation}
r(x|y)=\exp (-d(x,y))
\end{equation}
where $d(x,y)=|x-y|^2$ is the \emph{squared Euclidean distance}. Thus, maximizing $r(x|y)$ in eq.~\ref{eq.def-r} is equivalent to \emph{minimizing Euclidean distance}.

Our method takes as input the factual $y$ and the cluster model parameters. In order to account for plausibility, the user might specify a \emph{plausibility factor} $\epsilon$ to indicate whether the counterfactual should be located at the cluster boundary ($\epsilon=0$) or it should depart from the cluster boundary being placed inside the target cluster region for increasing values of $\epsilon>0$.  The cluster model along with the plausibility factor value $\epsilon$ define a \emph{constraint equation} that determines the set of points where the counterfactual should be located. We call this set \emph{constraint set} $CS_\epsilon$. If $\epsilon=0$, the constraint set is identical to the boundary $B$ between the two clusters.  

Our method also accounts for actionable and immutable features, i.e., which features are allowed to change and which are not. This is implemented by a user-defined binary mask vector $M$ that indicates which components of the parameter vector $z$ are allowed to change with respect to the factual $y$.  If $M_i=1$, $z_i$ is considered as free (actionable) parameter allowed to change. If $M_i =0$ then $z_i=y_i$, i.e. $z_i$ remains fixed (immutable). 

Given the factual $y$, the plausibility factor $\epsilon$ and the feature mask $M$, we present below how to compute counterfactuals in the case of $k$-means clustering as well as for Gaussian clusters with full, diagonal and spherical covariances. The solution is analytical in the $k$-means case and non-iterative, requiring the solution of a nonlinear  system with a single parameter, in the Gaussian case. 


\section{Counterfactuals for $k$-means clustering}
\label{sec:k-means}

Let $m_s$ and $m_t$ be two cluster centers. Based on the $k$-means cluster assignment rule, a point $z$ is assigned to the target cluster if
$|z-m_t|^2<|z-m_s|^2$. Therefore,
the cluster boundary contains those points $z$ for which $|z-m_s|^2=|z-m_t|^2$.  In order to account for plausibility, we define the constraint set $CS_\epsilon$ of candidate counterfactual locations, using the following constraint equation:
\begin{equation}
\label{eq:CSe-kmeans}
CS_e = \{z:  |z-m_s|^2 = |z-m_t|^2  + \epsilon |m_t-m_s|^2 \}, \epsilon \geq 0
\end{equation}
For $\epsilon=0$ the cluster boundary is defined.

We are given the factual $y$ and a binary mask $M$ indicating the elements of $z$ that are allowed to change during optimization. We proceed by splitting $z$ based on the free and fixed components: we denote by $z_f$ the subvector of free components of $z$ (with $M_i = 1$) and as $z_{fixed}$ the subvector of fixed components of $z$ (with $M_i = 0$), hence $z_{fixed} = y_{fixed}$. We wish to find the vector $z \in CS_e$ of the above form with minimum squared Euclidean distance from factual $y$.

Let $d_\epsilon = \epsilon |m_t - m_s|^2$.
The constraint equation for the set $CS_\epsilon$ can be written as 
\begin{equation}    
     z^\top (m_s - m_t) = \frac{|m_s|^2 - |m_t|^2 - d_\epsilon}{2}.
\end{equation}    
Let $v = m_s - m_t$ and $c = \frac{|m_s|^2 - |m_t|^2 - d_\epsilon}{2}$. 
Then the above constraint is written as $z^\top v = c$.

Let also $v_f$ and $v_{fixed}$  be the subvectors of $v$ with free and fixed components, respectively.
Then the constraint equation can be written as:
\begin{equation}
     z_f^\top v_f + y_{fixed}^\top v_{fixed} = c
\end{equation}
which is a linear constraint on $z_f$.

\begin{figure}[ht]
    \centering
    \includegraphics[width=0.65\linewidth]{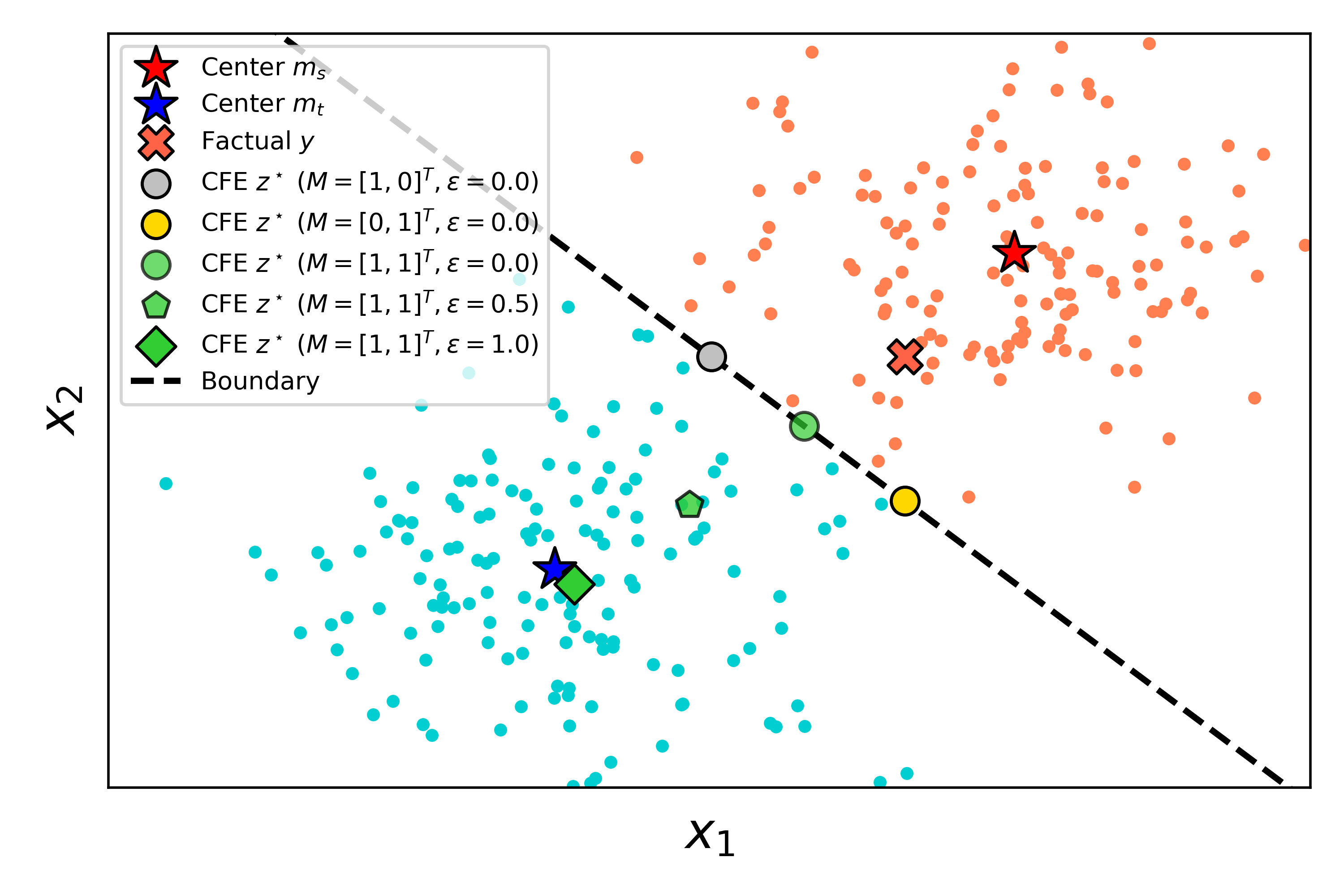}
    \caption{Illustration of the counterfactuals computed in the case of a 2-d synthetic dataset partitioned in two clusters using $k$-means. The factual $y$ is the same and the counterfactuals computed for several values of mask $M$ and palusibility factor $\epsilon$ are presented.}
    \label{fig:k-means-actionability}
\end{figure}

\noindent The optimization problem with parameter vector $z_f$ is defined as follows: 
\begin{center}
minimize  $|z_f - y_f|^2$ \\
subject to \\
$z_f^\top v_f = c - y_{fixed}^\top v_{fixed}$.
\end{center}

This is a quadratic minimization problem with a linear constraint, which can be solved either using Lagrange multipliers or by projecting $y_f$ onto the hyperplane defined by this constraint. The latter approach is straightforward and gives the solution for the free part of $z$:
\begin{equation}
\label{eq:kmeans-action}
     z_f^{\star} = y_f - \frac{(y_f^\top v_f - (c - y_{fixed}^\top v_{fixed}))}{|v_f|^2} v_f.
\end{equation}
The fixed components of $z^{\star}$ will be set equal to the corresponding $y_i$ values, i.e. $z^{\star}_{fixed}=y_{fixed}$.

It should be stressed that \emph{the above solution is valid only if} $v_f \neq 0$ or $y^\top_{fixed} v_{fixed} \neq c$. In the opposite case a valid counterfactual  does not exist.

Fig.~\ref{fig:k-means-actionability} provides a visual illustration of the optimal  counterfactuals computed by Eq.~\ref{eq:kmeans-action} for a 2-d synthetic dataset partitioned in two clusters using $k$-means. Several counterfactuals are presented for the same factual $y$ in order to illustrate the influence of the mask $M$ and the plausibility factor $\epsilon$. It is clear that, as $\epsilon$ increases, the counterfactual departs from the cluster boundary (where $\epsilon=0$) and moves inside the target cluster region. The influence of the actionability constraint is also illustrated. When $M=[1,0]$, only the horizontal coordinate is allowed to change, thus the counterfactual is placed in the same horizontal line with the factual $y$. When $M=[0,1]$, only the vertical coordinate is allowed to change, thus the counterfactual is placed in the same vertical line with the factual $y$.

\section{Counterfactuals for Gaussian clusters}
\label{sec:gaussian}
We now consider the case of Gaussian clusters starting with the more general case of full covariance matrices. 

\subsection{Gaussian clusters with full covariance}

Let $s$ and $t$ the source and target clusters with densities
$p_s(x) = \pi_s N(x; m_s, S_s)$ and $p_t(x) = \pi_t N(x; m_t, S_t)$  where $m_t$, $m_t$ are the mean vectors, $S_s$, $S_t$ are full covariance matrices and $\pi_s$, $\pi_t$ the cluster priors. 

\begin{figure}[ht]
    \centering
    \includegraphics[width=0.65\linewidth]{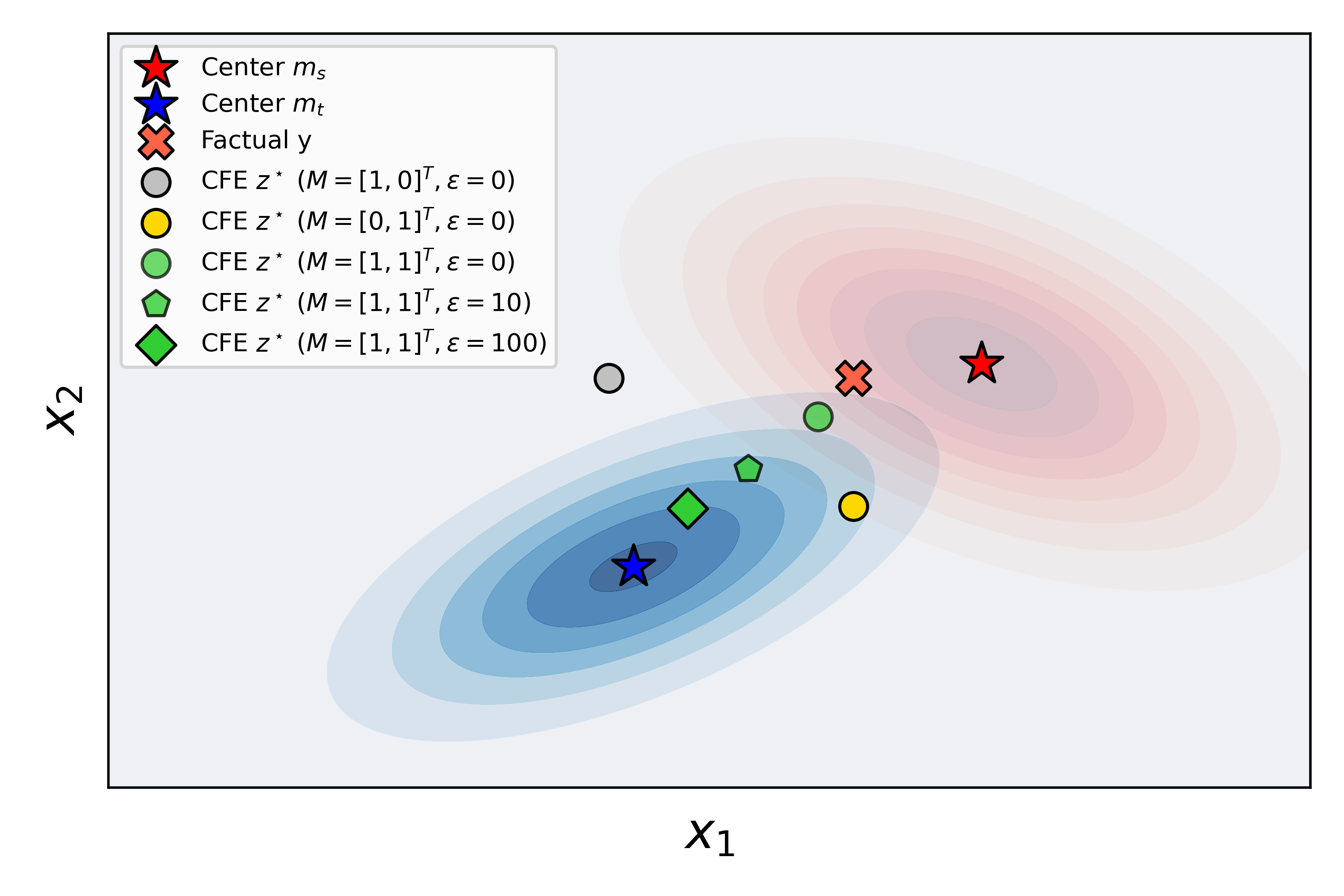}
    \caption{Illustration of the counterfactuals computed in the case of two Gaussian clusters with full covariances. The factual $y$ is the same and the counterfactuals computed for several values of mask $M$ and plausibility factor $\epsilon$ are presented.}
    \label{fig:gc-actionability}
\end{figure}

In probabilistic clustering, a point $z$ is assigned to the cluster of maximum density. Thus the cluster boundary contains the points $z$ with $p_s(z)=p_t(z)$. In order to account for plausibility, we define the constraint set $CS_\epsilon$ of possible counterfactual location, using the following constraint equation:
\begin{equation}
\label{eq:CSe}
CS_e = \{z: p_t(z) = (1+\epsilon) p_s(z) \}, \epsilon \geq 0
\end{equation}
For $\epsilon=0$ the cluster boundary is defined.

We are given the factual $y$, the mask vector $M$ and the plausibility factor $\epsilon$ and our goal is to find  $z \in CS_\epsilon$ that minimizes $| z - y |^2$ taking into account the mask  $M$. Based on mask $M$ we define the index sets:
\begin{itemize}
    \item Free indices: $F = \{i: M_i = 1\}$.
    \item Fixed indices: $G = \{i: M_i = 0\}$.
\end{itemize}
Therefore we can group  the indices of $z$ and $m_k$ as follows: 
\begin{itemize}
    \item $z= [z_F, z_G]^\top$, where $z_G = y_G$ (fixed).
    \item $m_k = [m_{kF}, m_{kG}]^\top$, for $k \in \{s, t \}$.
\end{itemize}
Moreover, the covariance inverses $S_k^{-1}$, $k \in \{s,t\}$ can also be partitioned into blocks:
\begin{equation*}
S_k^{-1} = \begin{bmatrix}
S_k^{-1,FF} & S_k^{-1,FG} \\
S_k^{-1,GF} & S_k^{-1,GG}
\end{bmatrix}
\end{equation*}

\noindent The optimization problem to be solved is defined: 

\begin{equation}
\text{minimize} \sum\limits_{i\in F} (z_i - y_i)^2 
\end{equation}
subject to the constraint:
\begin{align}
(z-m_t)^\top S_t^{-1}
(z -m_t) - (z-m_s)^\top S_s^{-1}(z-m_s) + \ln{\frac{|S_t|}{|S_s|} - 2\ln{\frac{\pi_t}{\pi_s}} + 2\ln (1 + \epsilon)} = 0
\end{align}
where the above equation is obtained by taking the logarithm of both sides in the constraint equation~\ref{eq:CSe}.

To simplify notation we define the constant 
\begin{equation}
\label{eq:A}
c_\alpha = \ln{\frac{|S_t|}{|S_s|} - 2\ln{\frac{\pi_t}{\pi_s}} + 2\ln (1 + \epsilon)}
\end{equation}
and replace in the above constraint equation.

The above quadratic optimization problem with a quadratic equality constraint can be solved by introducing a Lagrange multiplier $\lambda$ and defining the Lagrangian function:
\begin{align}
\mathcal{L}(z_F, \lambda) = \sum\limits_{i\in F} (z_i - y_i)^2 - \lambda \bigg[ (z-m_t)^\top S_t^{-1}(z-m_t)   - (z-m_s)^\top S_s^{-1}(z-m_s) + c_\alpha \bigg]
\end{align}
Taking the gradient of $L$ with respect to $z_F$ equal to zero we get
\begin{equation}
\label{eq:z-actionability-full}
    z_F = (I - \lambda D)^{-1}(y_F - \lambda d)
\end{equation}


\noindent where 
\begin{equation}
    D = S_t^{-1,FF} - S_s^{-1,FF}
\end{equation}
and 
\begin{align}
d = S_t^{-1,FF} m_{tF} - S_s^{-1,FF} m_{sF} - \big(S_t^{-1,FG}(z_G - m_{tG}) 
- S_s^{-1,FG}(z_G - m_{sG}) \big)
\end{align}


 


The constraint equation can be written as:
\begin{align}
\label{eq:constraint-actionability-full}
    & (z_F - m_{tF})^\top S_t^{-1, FF} (z_F - m_{tF}) + 2(z_F - m_{tF})^\top S_t^{-1, FG}(z_G - m_{tG}) + (z_{G} - m_{tG})^\top S_t^{-1, GG} (z_G - m_{tG}) \nonumber \\
    & - \big[ (z_F - m_{sF})^\top S_s^{-1, FF} (z_F - m_{sF})
    + 2(z_F - m_{sF})^\top S_s^{-1, FG}(z_G - m_{sG})  + (z_{G} - m_{sG})^\top S_s^{-1, GG} (z_G - m_{sG}) \big] \nonumber \\
    & + c_\alpha  = 0.
\end{align}

Substituting $z$ from Eq.~\ref{eq:z-actionability-full} into the constraint equation~\ref{eq:constraint-actionability-full},
the final equation to be solved for $\lambda$ becomes:
\begin{align}
\label{eq:lambda-actionability-full}
     (y_F - \lambda d)^\top (I - \lambda D)^{-1} D  (I - \lambda D)^{-1} (y_F - \lambda d) - 2(y_F - \lambda d)^{\top}(I - \lambda D)^{-1}e + c_f + c_\ell + c_g + c_\alpha = 0
\end{align}
where 
\begin{equation}
    e = S_t^{-1, FF}m_{tF} -  S_s^{-1, FF}m_{sF}
\end{equation}
\begin{equation}
    c_f = m_{tF}^\top S_t^{-1, FF} m_{tF} - m_{sF}^\top S_s^{-1, FF} m_{sF}
\end{equation}
\begin{align}
    c_g  = (z_G - m_{tG})^\top S^{-1, GG}_t (z_G - m_{tG}) - (z_G - m_{sG})^\top S^{-1, GG}_s (z_G - m_{sG})
\end{align}
\begin{align}
    c_\ell  = 2(z_F - m_{tF})^\top S_t^{-1,FG}(z_G - m_{tG}) - 2(z_F - m_{sF})^\top S_s^{-1,FG}(z_G - m_{sG})
\end{align}

Once the solution $\lambda^{\star}$ is found, it is substituted in Eq.~\ref{eq:z-actionability-full} to provide the counterfactual $z_F^{\star}$. For the fixed parameters it holds that $z^{\star}_G = y_G$.

In what concerns the uniqueness of the solution, if matrix $S_t^{-1, FF} - S_s^{-1, FF}$ is positive or negative definite then a unique solution exists. Otherwise, there is possibility for one, multiple or no solutions.

Fig.~\ref{fig:gc-actionability} provides a visual illustration of the counterfactuals computed for the case of two Gaussian clusters with full covariances. It should be noted that the cluster boundary is quadratic in this case. 
The figure shows multiple counterfactuals corresponding to the same factual instance $y$ highlighting their behavior as $M$ and $\epsilon$ varies.
As $\epsilon$ increases, the counterfactuals gradually move away from the cluster boundary (where $\epsilon=0$) and further into the interior of the target cluster. The impact of the actionability constraint is also demonstrated. When $M=[1,0]$, only the horizontal coordinate is allowed to vary, resulting in the counterfactual being horizontally aligned with the factual point $y$. In contrast, when $M=[0,1]$, only the vertical coordinate is adjustable, placing the counterfactual on the same vertical line as $y$.

\subsection{Gaussian clusters with diagonal covariance}

Let $p_s(x) = \pi_s N(x; m_s, S_s)$ and $p_t(x) = \pi_t N(x; m_t, S_t)$  where $S_s$ and $S_t$ are \emph{diagonal} covariance matrices, i.e. \begin{center}
	$S_s = \text{diag}(\sigma_{s1}^2,\ldots, \sigma_{sd}^2)$ 
	and $S_t = \text{diag}(\sigma_{t1}^2,\ldots, \sigma_{td}^2)$. 
\end{center}
The diagonal nature of the covariances allows for component-wise computations and formulas compared to the more complicated full covariance case.


The constraint \( p_t(z) = (1+\epsilon) \cdot p_s(z) \) translates to:
\begin{align}
\label{eq:constraint-diagonal-actionability}
 \sum_{i \in F} \left( \frac{(z_i - m_{ti})^2}{\sigma^2_{ti}} - \frac{(z_i - m_{si})^2}{\sigma^2_{si}} \right) +  \sum_{i \in G} \left( \frac{(y_i - m_{ti})^2}{\sigma^2_{ti}} - \frac{(y_i - m_{si})^2}{\sigma^2_{si}} \right) + c_\alpha =0
\end{align}
where $c_\alpha$ is given by Eq.~\ref{eq:A} with $|S_k|=\prod_{j=1}^d \sigma^2_{kj}$, $k\in \{s,t\}$.

Our objective is to minimize $\sum_{i \in F} (z_i - y_i)^2 $
subject to the constraint equation~\ref{eq:constraint-diagonal-actionability}.
We define the Lagrangian:
\begin{align}
\mathcal{L}(z, \lambda) = \sum_{i \in F} (z_i - y_i)^2 + \lambda \left( \sum_{i \in F} \left( \frac{(z_i - m_{ti})^2}{\sigma^2_{ti}} - \frac{(z_i - m_{si})^2}{\sigma^2_{si}} \right) + c_h + c_\alpha \right)
\end{align}
where $c_h = \sum_{i \in G} \left( \frac{(y_i - m_{ti})^2}{\sigma^2_{ti}} - \frac{(y_i - m_{si})^2}{\sigma^2_{si}} \right)$ is a constant. For $i \in F$, taking the partial derivative of $\mathcal{L}$ with respect to $z_i$ and setting to zero we get:

\begin{equation}
\label{eq:z-diagonal-actionability}
z_i = \frac{y_i + \lambda \left( \frac{m_{si}}{\sigma^2_{si}} - \frac{m_{ti}}{\sigma^2_{ti}} \right)}{1 + \lambda \left( \frac{1}{\sigma^2_{ti}} - \frac{1}{\sigma^2_{si}} \right)}.
\end{equation}

The partial derivative of the Lagrangian with respect to $\lambda$ gives the constraint equation:
\begin{equation}
\sum_{i \in F} \left( \frac{(z_i - m_{ti})^2}{\sigma^2_{ti}} - \frac{(z_i - m_{si})^2}{\sigma^2_{si}} \right) + c_h + c_\alpha =0.
\end{equation}

Substituting $z_i$ from equation~\ref{eq:z-diagonal-actionability} into the above constraint equation we get an equation that contains only the $\lambda$ parameter:
\begin{align}
\label{eq:lambda-actionability-diagonal}	
\sum_{i \in F} \frac{\left( \frac{y_i - \lambda \left( \frac{m_{si}}{\sigma^2_{si}} - \frac{m_{ti}}{\sigma^2_{ti}} \right)}{1 + \lambda \left( \frac{1}{\sigma^2_{si}} - \frac{1}{\sigma^2_{ti}} \right)} - m_{ti} \right)^2}{\sigma^2_{si}}  - \sum_{i \in F}\frac{\left( \frac{y_i + \lambda \left( \frac{m_{si}}{\sigma^2_{si}} - \frac{m_{ti}}{\sigma^2_{ti}} \right)}{1 + \lambda \left( \frac{1}{\sigma^2_{si}} - \frac{1}{\sigma^2_{ti}} \right)} - m_{si} \right)^2}{\sigma^2_{ti}} + c_h + c_\alpha = 0.
\end{align}

This equation can be solved using numerical methods. If a solution $\lambda^{\star}$ is found, then each free $z_i$ can be computed by substituting $\lambda^{\star}$ in Eq.~\ref{eq:z-diagonal-actionability}.
The fixed elements are simply $z_i^{\star} = y_i$ for $i \in G$.

In what concerns the uniqueness of the solution, it is guaranteed in the case where the sign of $\frac{1}{\sigma_{si}^2} - \frac{1}{\sigma_{ti}^2}$ is the same for all free dimensions $i\in F$. Otherwise, there is possibility for one, multiple or no solutions.

\subsection{Gaussian clusters with spherical covariance}

Let $p_s(x) = \pi_s N(x; m_s, S_s)$ and $p_t(x) = \pi_t N(x; m_t, S_t)$  where $S_s$ and $S_t$ are spherical covariance matrices \begin{center}
	$S_s = \sigma_s^2 I \quad \text{and} \quad S_t = \sigma_t^2 I$. 
\end{center}


The constraint $p_t(z) = (1+\epsilon) \cdot p_s(z)$ translates to:
\begin{equation}
\label{eq:spherical_constraint}
\frac{| z - m_t |^2}{\sigma_t^2} - \frac{| z - m_s |^2}{\sigma_s^2} + c_\alpha =0.
\end{equation}
where $A$ is given by Eq.~\ref{eq:A} with $|S_k|=\sigma^{2d}_{k}$, $k\in \{s,t\}$.


\noindent The optimization problem is to minimize with respect to free $z_i$ the objective $\sum_{i \in F} (z_i - y_i)^2$,
subject to the constraint of Eq.~\ref{eq:spherical_constraint}

The solution to this problem can be obtained as a special case of the previously presented equations for the diagonal covariance case. By replacing $\sigma_{sj}^2 = \sigma_s^2$ and $\sigma_{tj}^2 = \sigma_t^2$ for all $j$ in Eq.~\ref{eq:z-diagonal-actionability} and Eq.~\ref{eq:lambda-actionability-diagonal}, we find the following equations for $z_i \in F$:

\begin{equation}
\label{eq:z-spherical-actionability}		
z_i = \frac{y_i - \lambda \left( \frac{m_{si}}{\sigma_s^2} - \frac{m_{ti}}{\sigma_t^2} \right)}{1 + \lambda \left( \frac{1}{\sigma_t^2} - \frac{1}{\sigma_s^2} \right)}
\end{equation}

and for $\lambda$:
\begin{align}
& \frac{1}{\sigma_t^2} \left( \sum_{i \in F} \left( \frac{y_i - m_{ti} - \lambda \left( \frac{m_{si}}{\sigma_s^2} - \frac{m_{ti}}{\sigma_t^2} \right)}{1 + \lambda \left( \frac{1}{\sigma_t^2} - \frac{1}{\sigma_s^2} \right)} \right)^2 + \sum_{i \in G} (y_i - m_{ti})^2 \right) - \nonumber \\ 
& \frac{1}{\sigma_s^2} \left( \sum_{i \in F} \left( \frac{y_i - m_{si} - \lambda \left( \frac{m_{si}}{\sigma_s^2} - \frac{m_{ti}}{\sigma_t^2} \right)}{1 + \lambda \left( \frac{1}{\sigma_t^2} - \frac{1}{\sigma_s^2} \right)} \right)^2 + \sum_{i \in G} (y_i - m_{si})^2 \right) \nonumber \\ & + c_\alpha = 0.
\end{align}

This equation is nonlinear in $\lambda$ and is numerically solved. Once the solution $\lambda^\star$ is found, it can be substituted back into Eq.~\ref{eq:z-spherical-actionability} to find the optimal values of the free parameters $z^{\star}_i$ ($i \in F$).
For the fixed parameters $i \in G$ it holds that $z^{\star}_i = y_i$. It should be noted that a unique solution exists in this case.

\section{Experimental Results}
\label{sec:experiments}
To demonstrate the effectiveness of our approach
called CounterFactuals for Clustering ({\cfclust}), at first we present an illustrative example with digit images and then we provide 
performance results.

When using our method for Gaussian clustering, to solve for $\lambda$, we use the root finding method of SciPy library\footnote{\url{https://docs.scipy.org/doc/scipy/reference/optimize.root-hybr.html}}.

\subsection{An illustrative example}

We used well-studied OptDigits~\cite{kelly2024uci}, a 10-class dataset which contains grayscale image of handwritten digits. Each digit image is represented by a 64-dimensional feature vector. These features correspond to the intensity values of an $8\times8$ grid of pixels. We ignored the class labels and applied $k$-means to partition the dataset into ten clusters. Then we designated the cluster with the majority of images representing the digit zero as the source cluster $C_s$ and the cluster with the majority of images representing the digit two as the target cluster $C_t$.

\begin{figure}[ht]
\centering
\resizebox{0.65\textwidth}{!}{ 
  \begin{minipage}{\textwidth} 
  \centering
    \captionsetup[subfloat]{labelformat=empty}

    \subfloat[(a) center $m_s$]{
    \includegraphics[width=0.2\linewidth]{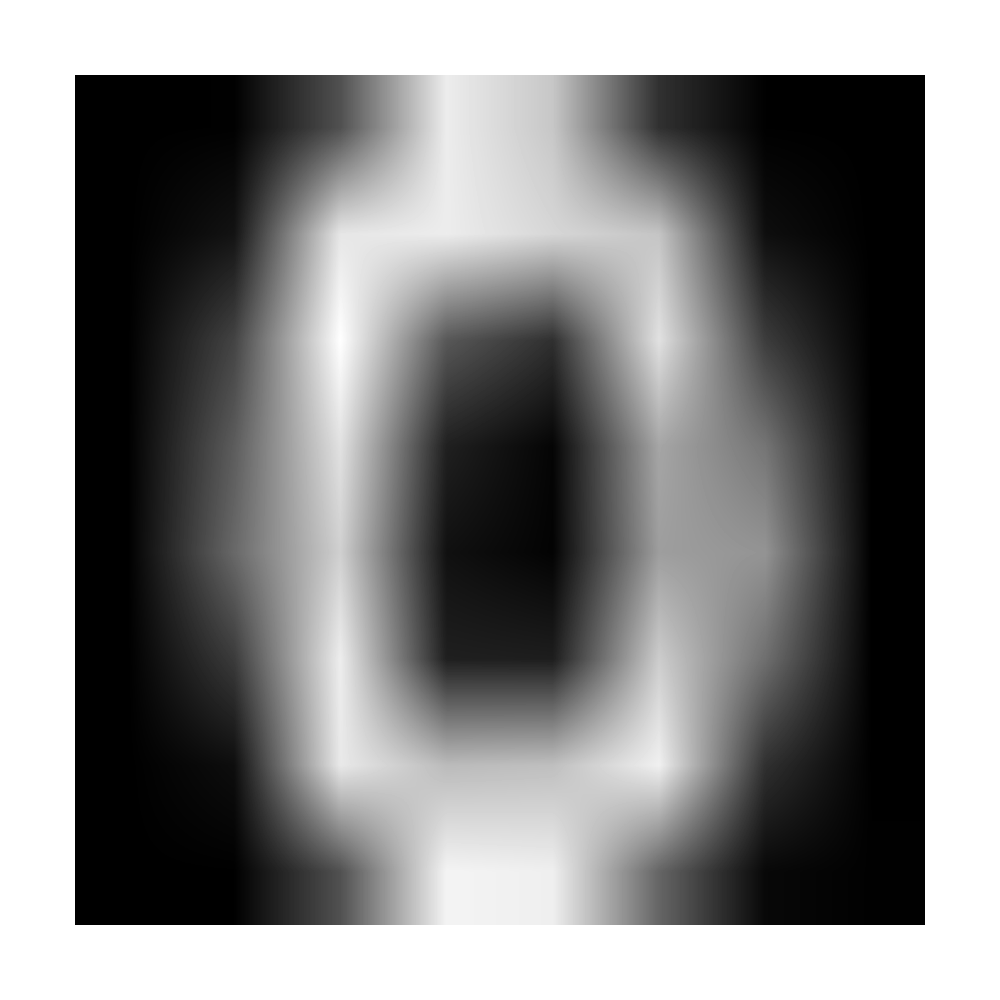}
    }
    \subfloat[(b) center $m_t$]{
    \includegraphics[width=0.2\linewidth]{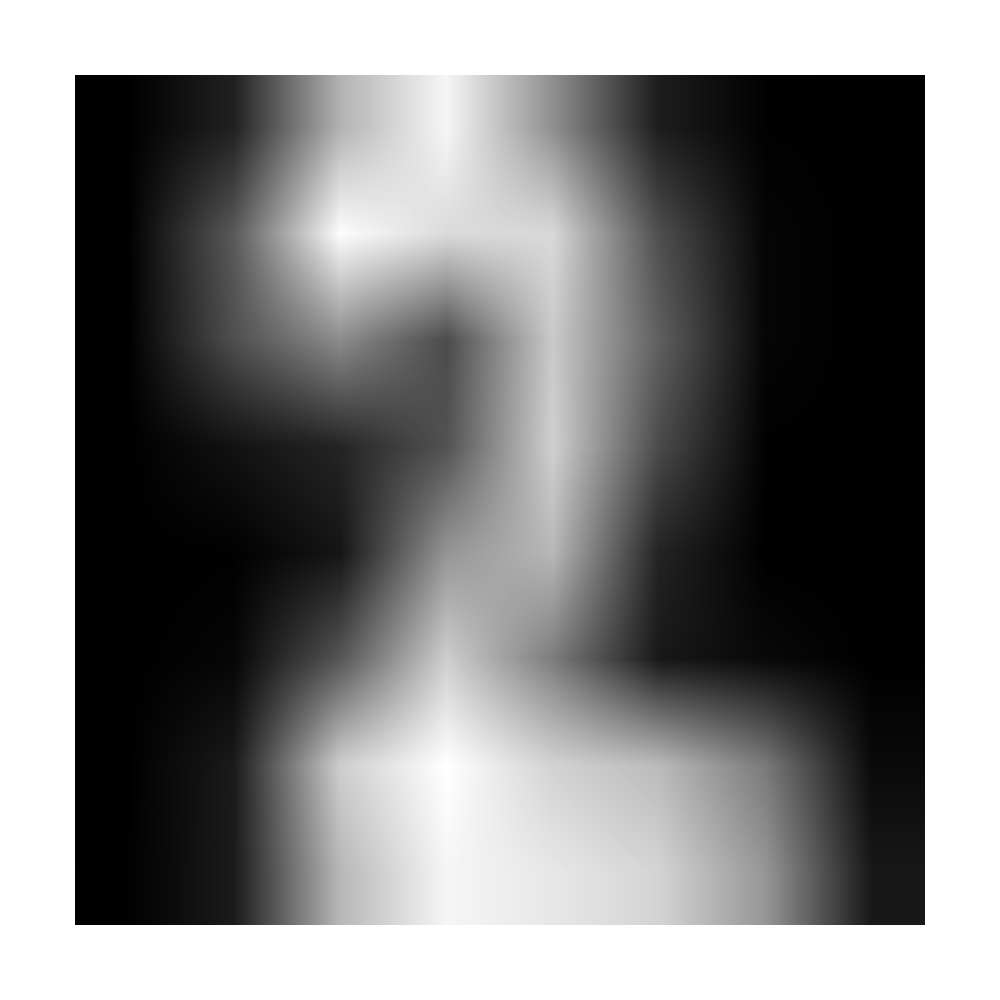}
    }
    \subfloat[(c) factual $y$]{
    \includegraphics[width=0.2\linewidth]{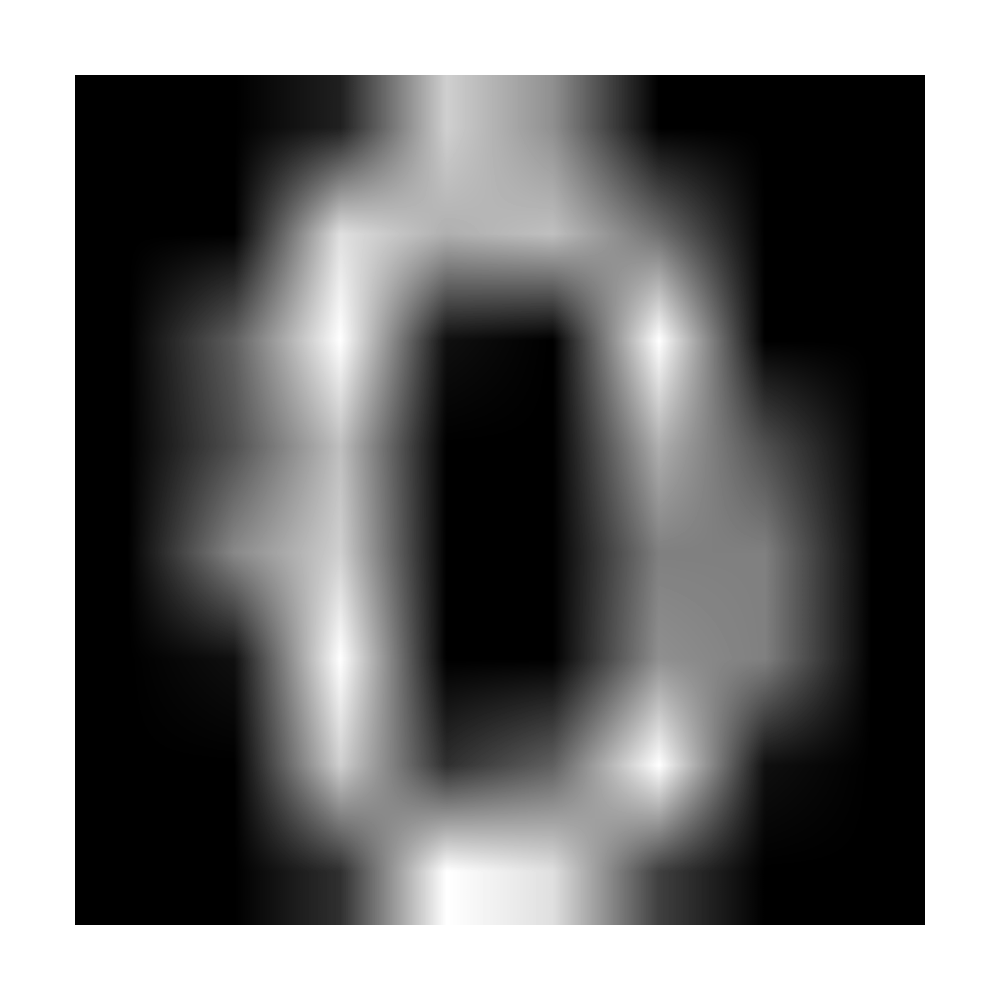}
    }

    \subfloat[(d) $\epsilon=0.0$]{
    \includegraphics[width=0.2\linewidth]{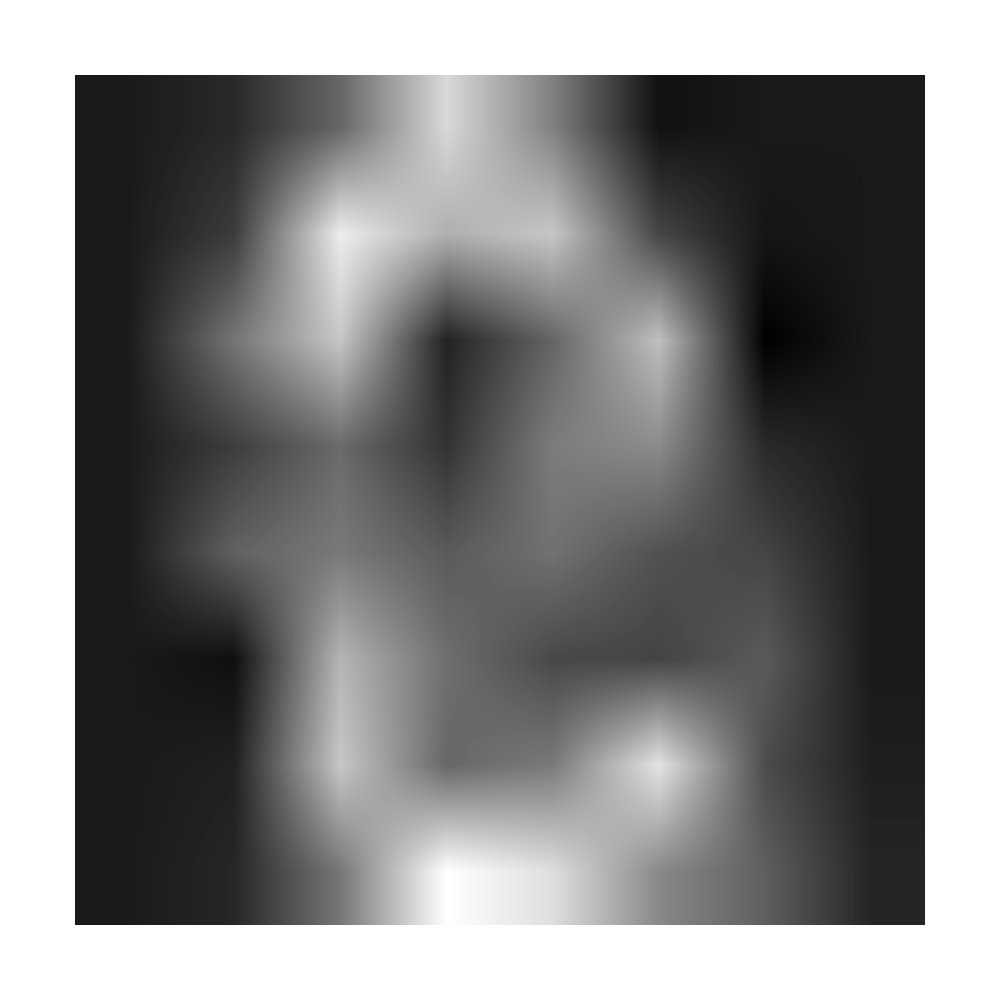}
    }
    \subfloat[(e) $\epsilon=0.33$]{
    \includegraphics[width=0.2\linewidth]{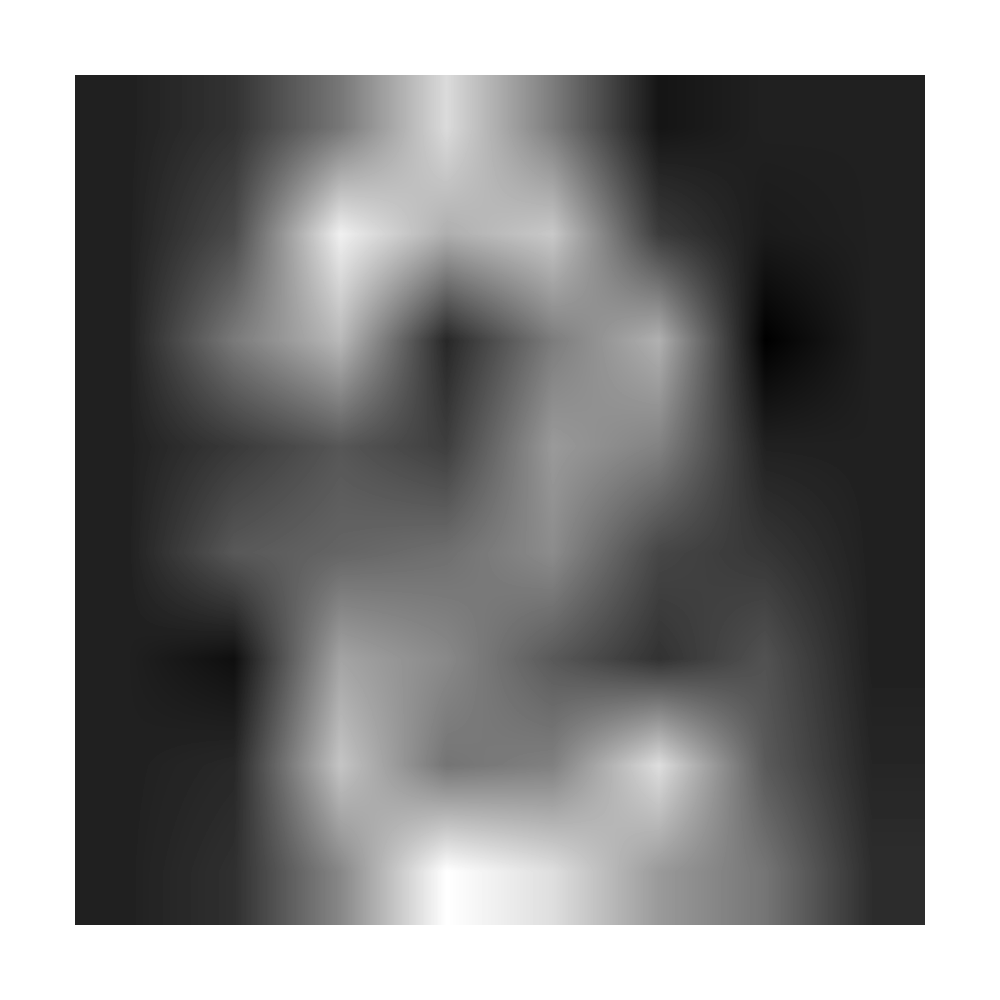}
    }
    \vspace{0.5em}
    \subfloat[(f) $\epsilon=0.66$]{
    \includegraphics[width=0.2\linewidth]{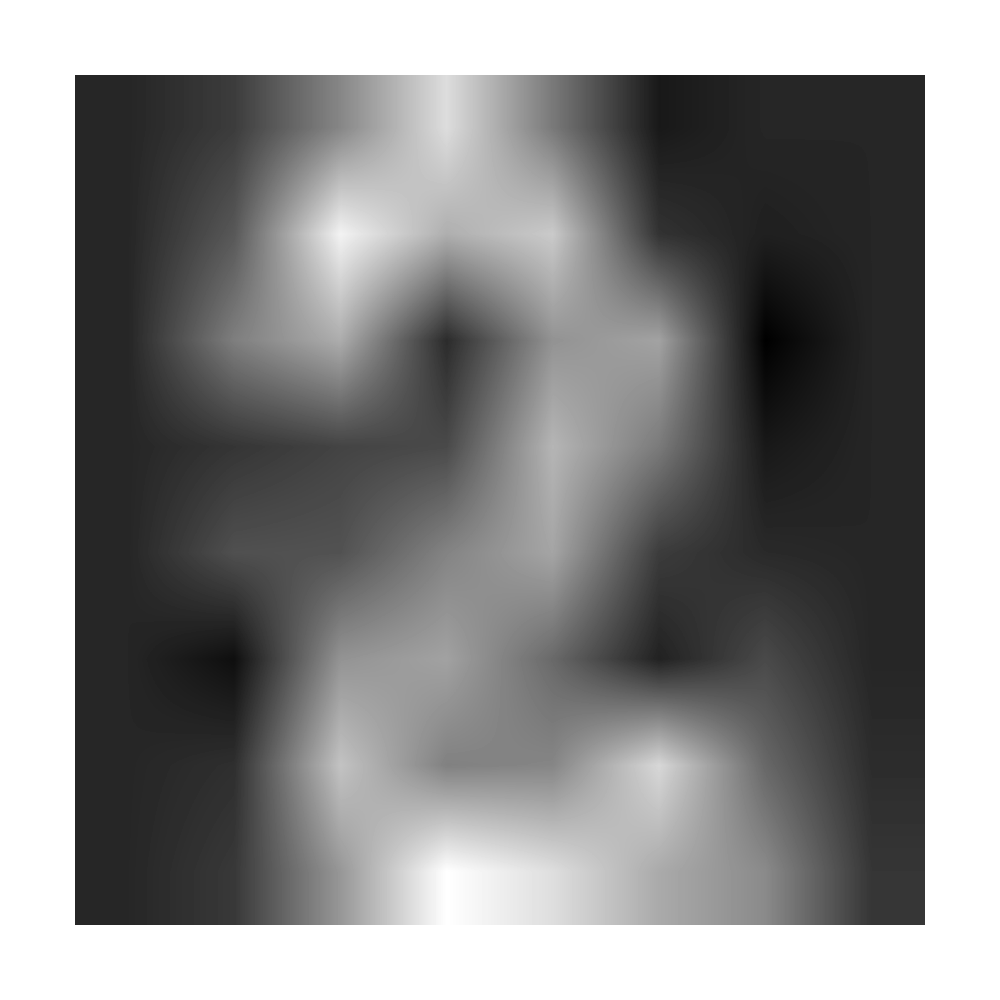}
    }
    \subfloat[(g) $\epsilon=1.0$]{
    \includegraphics[width=0.2\linewidth]{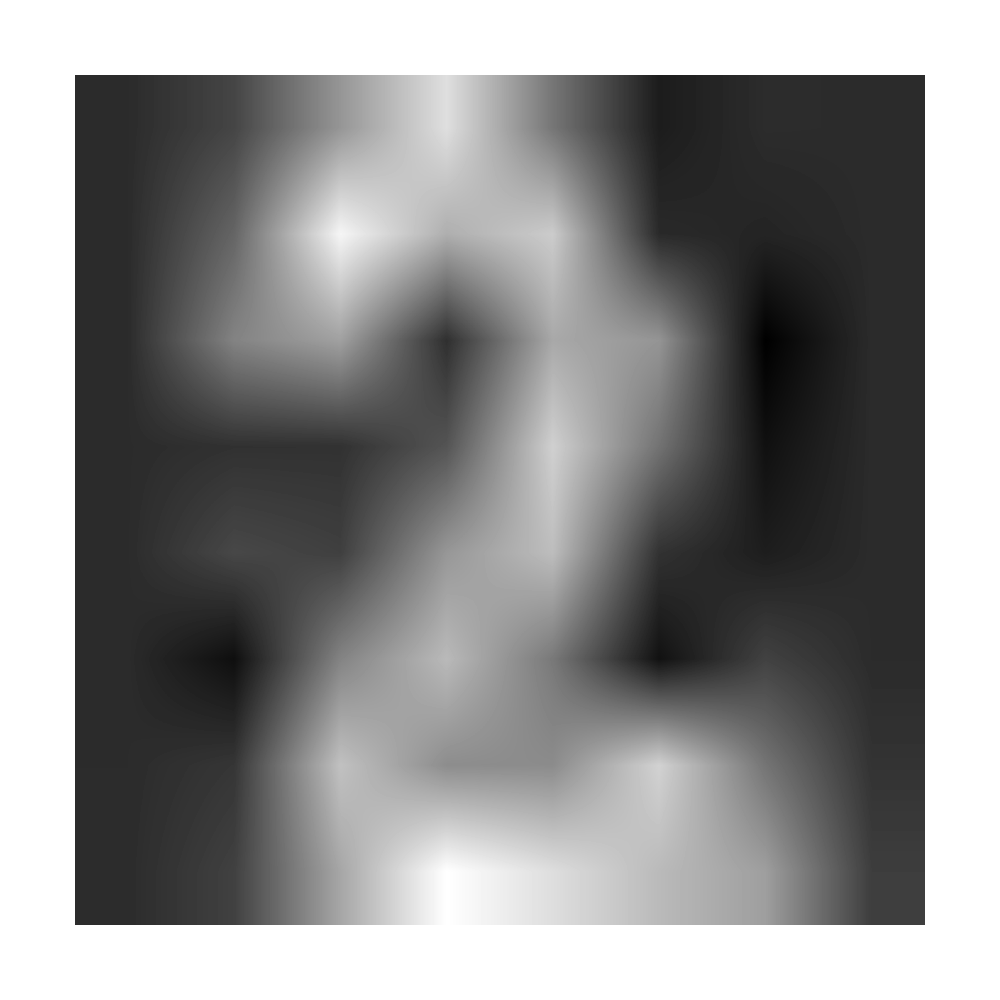}
    }

    \subfloat[(h) $\epsilon=0.0$]{
    \includegraphics[width=0.2\linewidth]{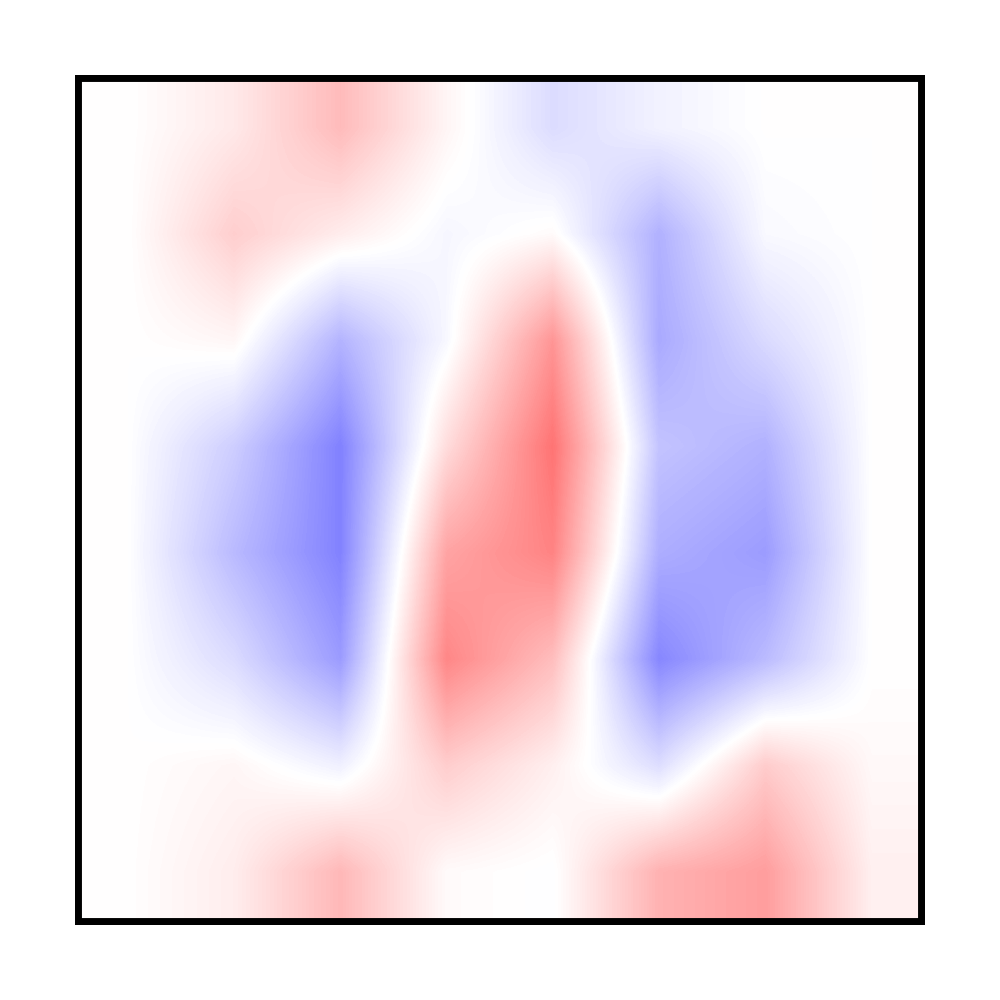}
    }
    \subfloat[(i) $\epsilon=0.33$]{
    \includegraphics[width=0.2\linewidth]{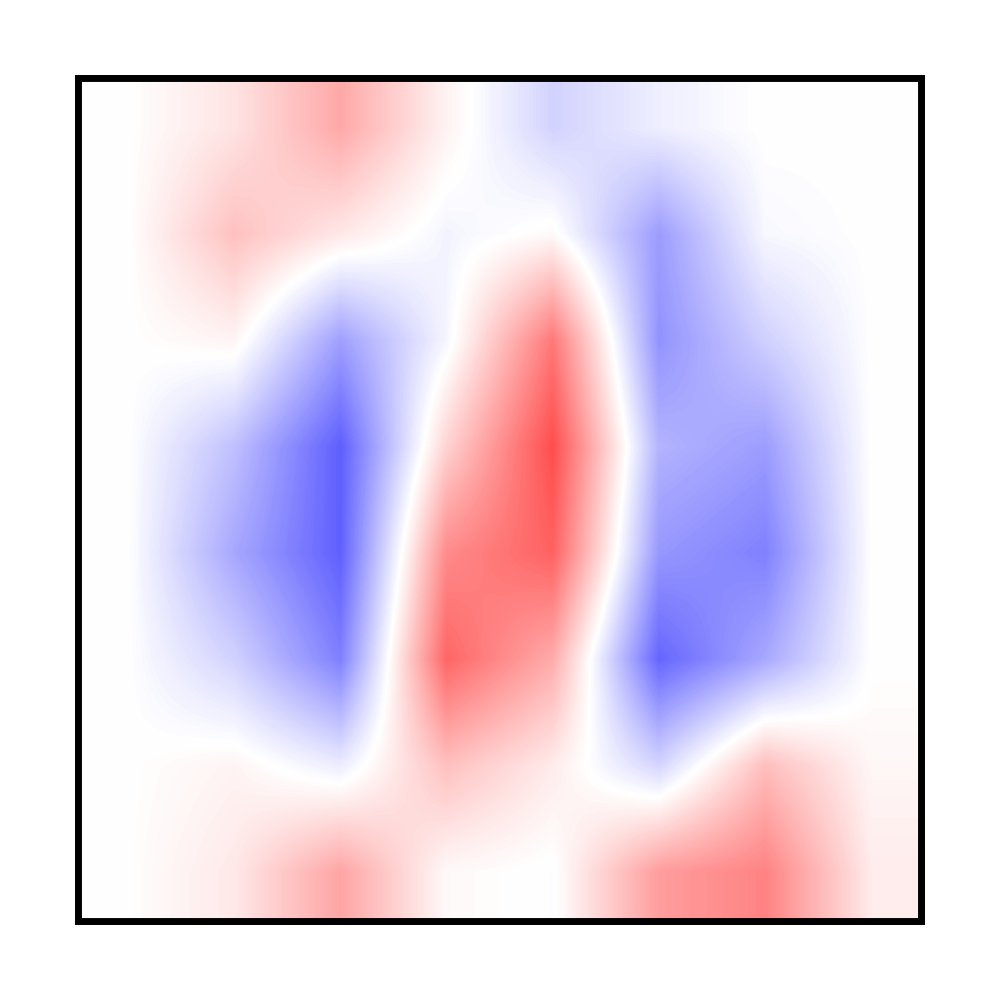}
    }
    \vspace{0.5em}
    \subfloat[(j) $\epsilon=0.66$]{
    \includegraphics[width=0.2\linewidth]{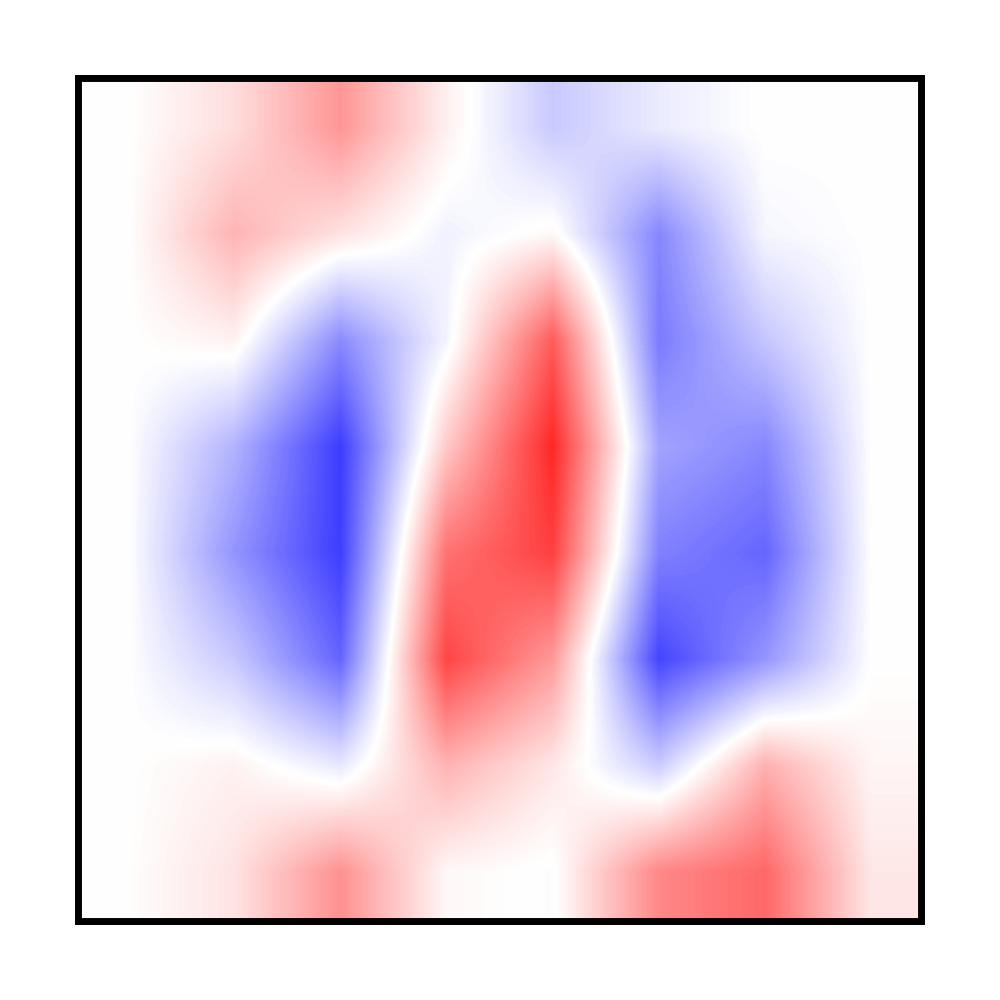}
    }
    \subfloat[(k) $\epsilon=1.0$]{
    \includegraphics[width=0.2\linewidth]{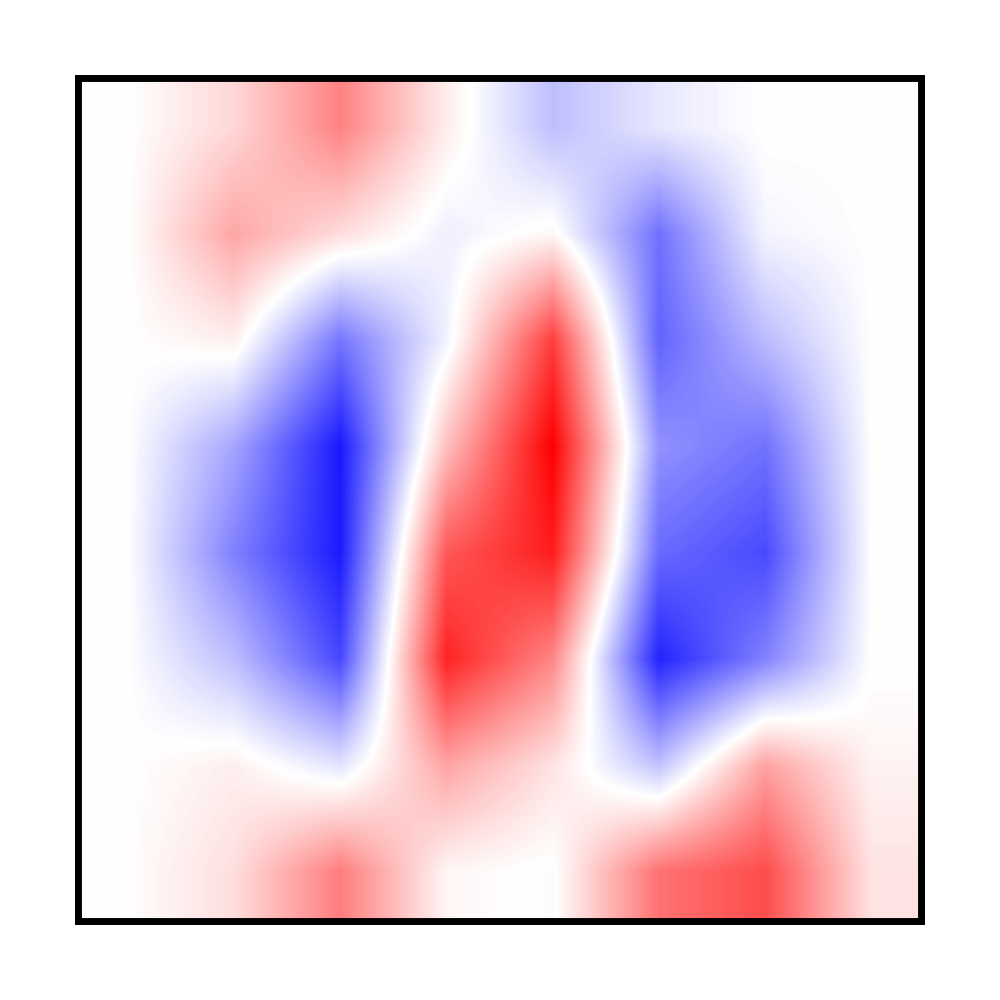}
    }
  \end{minipage}
}
\caption{Illustrative example using images of the OptDigits dataset. Top row: images of the two cluster centers and the factual. Middle row: counterfactual images generated by our method for increasing values of plausibility factor $\epsilon$. Bottom row: visualization the pixel-wise differences between the corresponding counterfactual image (shown in the middle row) and the factual image. Blue and red colors indicate negative and positive differences respectively.}
\label{fig:digits}
\end{figure}

The top row of Fig.~\ref{fig:digits} presents the images of the centers $m_s$ and $m_t$ of the two clusters. It also presents the image of the factual instance $y$ that belongs to the source cluster. Given the 64-dimensional vectors $m_s$, $m_t$ and $y$, we used Eq.~\ref{eq:kmeans-action} assuming no immutable features  ($M_i=1$, for all $i$) to compute the counterfactuals for increasing values of the plausibility factor $\epsilon$. The images of the generated counterfactuals are presented in the middle row of the figure. It can be observed that when $\epsilon=0$, the shape of counterfactual digit is actually a blend of zero and two, since the counterfactual lies on the cluster boundary. As the value of $\epsilon$ increases, the shape of the counterfactual digit clearly corresponds to digit two. 

Each image in the third row of the figure visualizes the pixel-wise differences between the corresponding counterfactual image (shown in the middle row) and the factual image. Red colors indicate positive differences, meaning that in order to generate the counterfactual the corresponding pixels should 'turn on' (their intensity should increase compared to the factual). Blue colors indicate negative differences, meaning that in order to the corresponding pixels should 'turn off' (their intensity should decrease compared to the factual).
In this way, a visual explanation is directly provided of how to modify the factual in order to produce the counterfactual. It is interesting to note our generation method, correctly operating towards minimum changes, does not affect the values of the pixels lying on the two vertical borders of the factual image.

\subsection{Evaluation}
In this set of experiments, we provide comparative experimental results of the performance of {\cfclust} using synthetic and real datasets.

\subsubsection{Experimental Setup}
For each dataset considered, we first apply clustering using either $k$-means or by training a Gaussian mixture model with full covariance. Then we select a source cluster $C_s$ and a target cluster $C_t$ and proceed with the evaluation of the counterfactual generation methods. 

Since there is no previous research on generating counterfactuals for clustering, 
we compare our approach with previous research on generating counterfactuals for classification as follows.
We  first build a binary classifier with  class labels corresponding to the  source and target clusters.
Given an example $y$ in $C_s$, we apply a known counterfactual generation algorithm for classification to produce a counterfactual $z$.
To avoid classification errors, instead of using the cluster labels to train the classifier, in the case of the  $k$-means clustering, we use  a \emph{Logistic Regression} classifier (LR) whose decision hyperplane coincides with the linear cluster boundary defined by the cluster centers $m_s$ and $m_t$. 
In the Gaussian clustering case, the cluster boundary is quadratic and we specify as our classifier, a  
\emph{Quadratic Discriminant Analysis} (QDA) classifier whose parameters are determined by the means, covariances and priors of the Gaussian clusters.


For generating counterfactuals in the above classification framework,  we employ two state-of-the-art algorithms, namely  DiCE ({\dice}) \cite{mothilal2020explaining} and GuidedByPrototypes ({\prt}) \cite{van2021interpretable}. 
For {\textsc{dice}}, we use the implementation provided by the authors\footnote{\url{https://github.com/interpretml/DiCE}}, while for {\textsc{prt}}, we use the  implementation in Alibi Explain\footnote{\url{https://docs.seldon.io/projects/alibi/en/stable/}}.
Their parameters determined after careful tuning are described in the Appendix. 
{\dice} and {\prt}  do not offer a parameter for directly adjusting plausibility as {\cfclust} does. We set $\epsilon=10^{-5}$, thus finding counterfactuals very close to the cluster boundary.

We evaluate our method using two synthetic and three real datasets.
Two synthetic datasets were generated using the scikit-learn $make\_blobs$\footnote{\url{https://scikit-learn.org/stable/modules/generated/sklearn.datasets.make_blobs.html}} function: one with two features and two Gaussian clusters ({\two}) and one with three features and two Gaussian clusters ({\three})  
We have also used three well-known real datasets, \textit{{\Iris}}, \textit{\Wine} and {\textit{Pendigits}} \cite{kelly2024uci}.
The {\Iris} dataset has 3 categories corresponding to three species of the Iris flower and 4 numerical features describing morphological properties of flowers.
The {\Wine} dataset has 3 categories of wine types and 13 numerical features describing wine chemical properties.
The {\pendigits} dataset comprises 10 digit categories and 16 numerical features defining the pen's trajectory. Note that class labels were not used by the clustering algorithms and the number of clusters was set equal to the number of classes.


\subsubsection{Experimental Results}
We present results for all datasets, considering both all features actionable as well as some of them immutable. 
For each case we generate counterfactuals for 50 factuals randomly selected  from the source cluster.

In contrast to {\cfclust}, {\dice} and {\prt} do not always generate instances that belong to the target cluster.
The reported distance results concern only those factuals for which all methods generated solutions in the target cluster. 
In Table \ref{tab:success-lr-qda} in the Appendix, we provide the percentage of factuals for which the generated instance belongs to the target cluster. 

\begin{figure}[h!]
    \centering
    \resizebox{0.9\textwidth}{!}{ 
        \begin{minipage}{\textwidth} 
        \centering
            \begin{subfigure}[b]{0.3\textwidth}
                \centering
                \includegraphics[width=\textwidth]{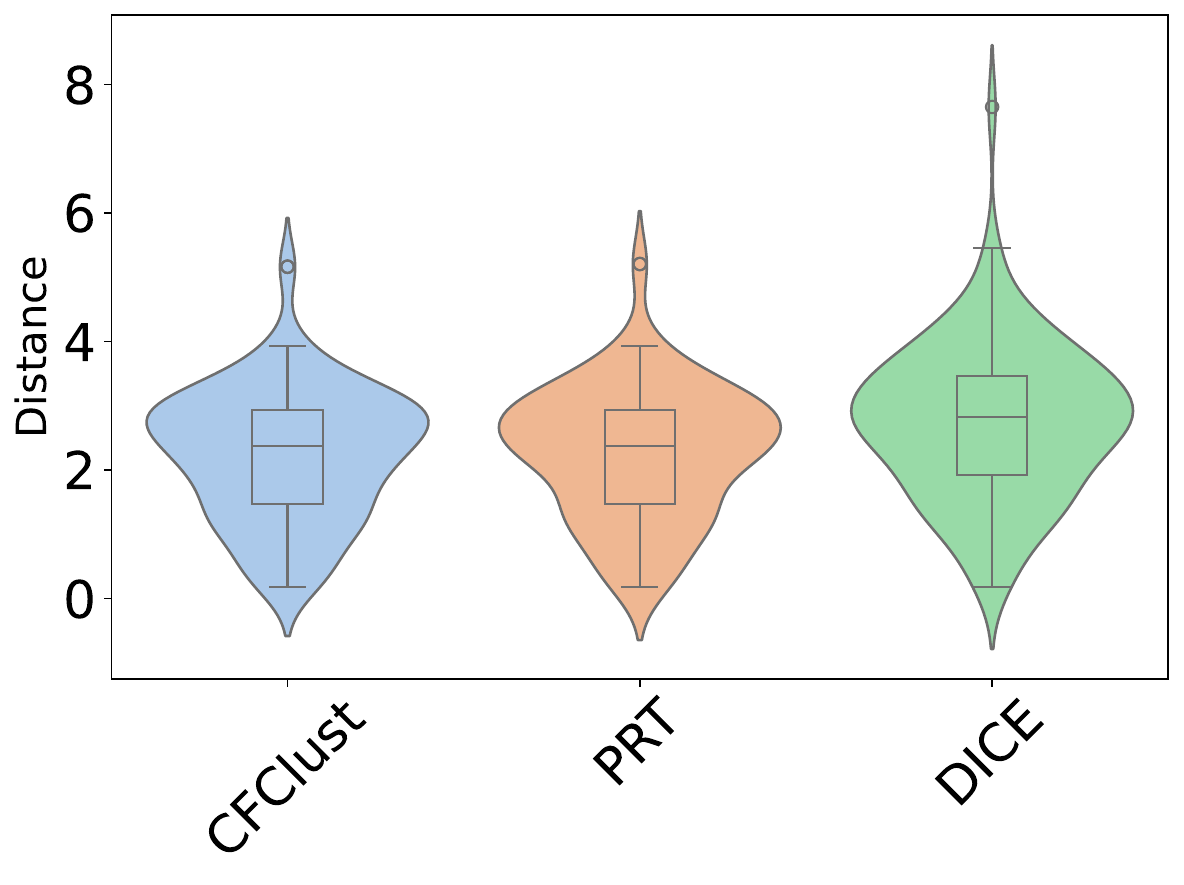} 
                \label{fig:2d-lr-no}
            \end{subfigure}
            \begin{subfigure}[b]{0.3\textwidth}
                \centering
                \includegraphics[width=\textwidth]{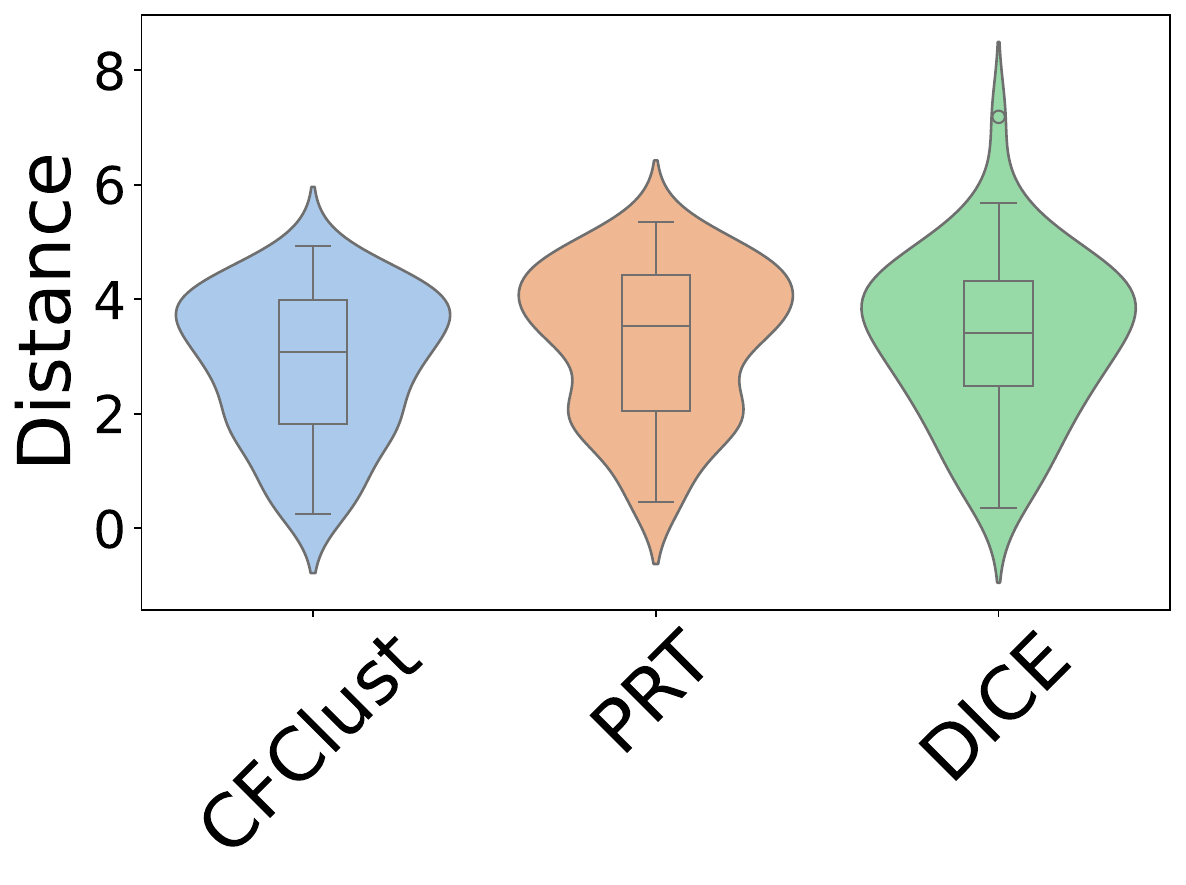} 
                \label{fig:2d-lr-imm1}
            \end{subfigure}
            \begin{subfigure}[b]{0.3\textwidth}
                \centering
                \includegraphics[width=\textwidth]{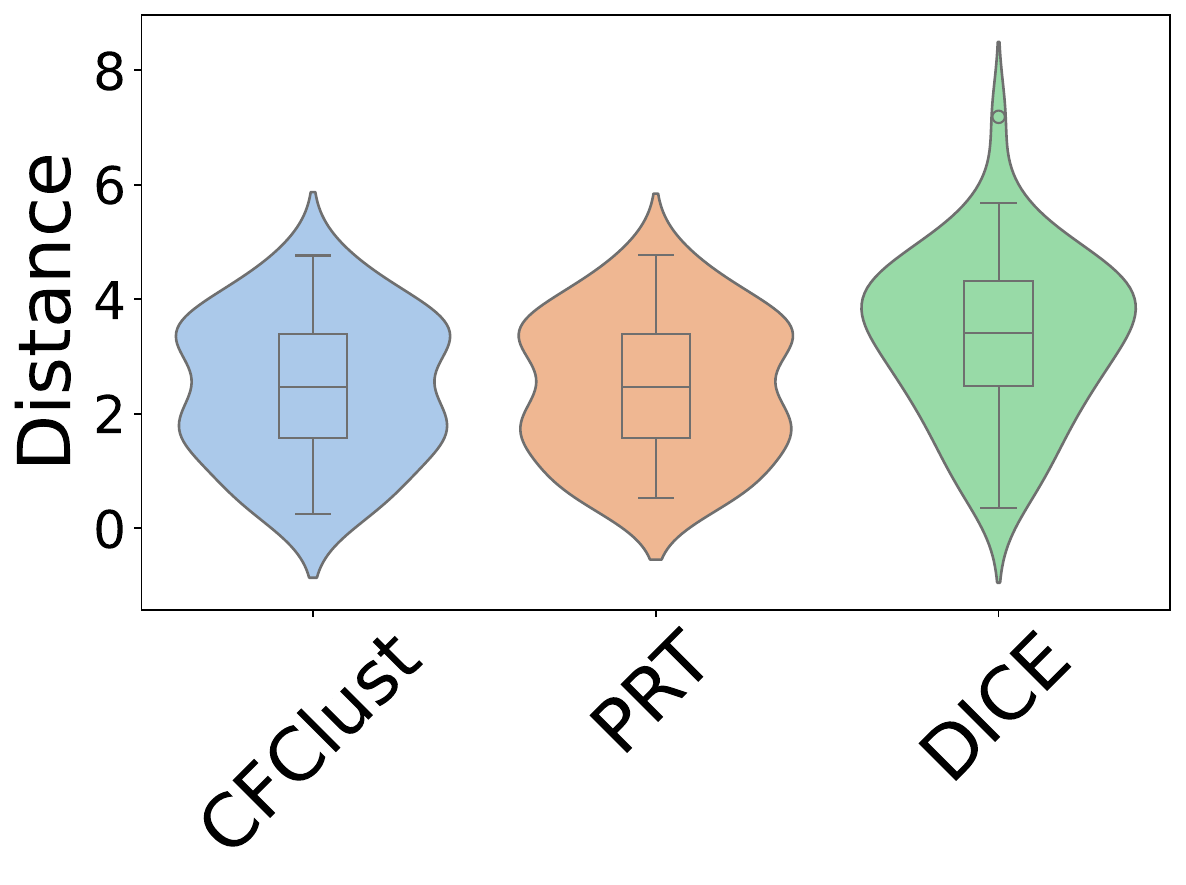} 
                \label{fig:2d-imm2-lr}
            \end{subfigure}
            \hfill
        \end{minipage}
    }
    \caption{$k$-means for {\two}: (left) no immutable features, (middle)-(right) one immutable feature.}
    \label{fig:lr-2d}
\end{figure}

\begin{figure}[h!]
    \centering
    \resizebox{0.9\textwidth}{!}{ 
        \begin{minipage}{\textwidth} 
        \centering
            \begin{subfigure}[b]{0.3\textwidth}
                \centering
                \includegraphics[width=\textwidth]{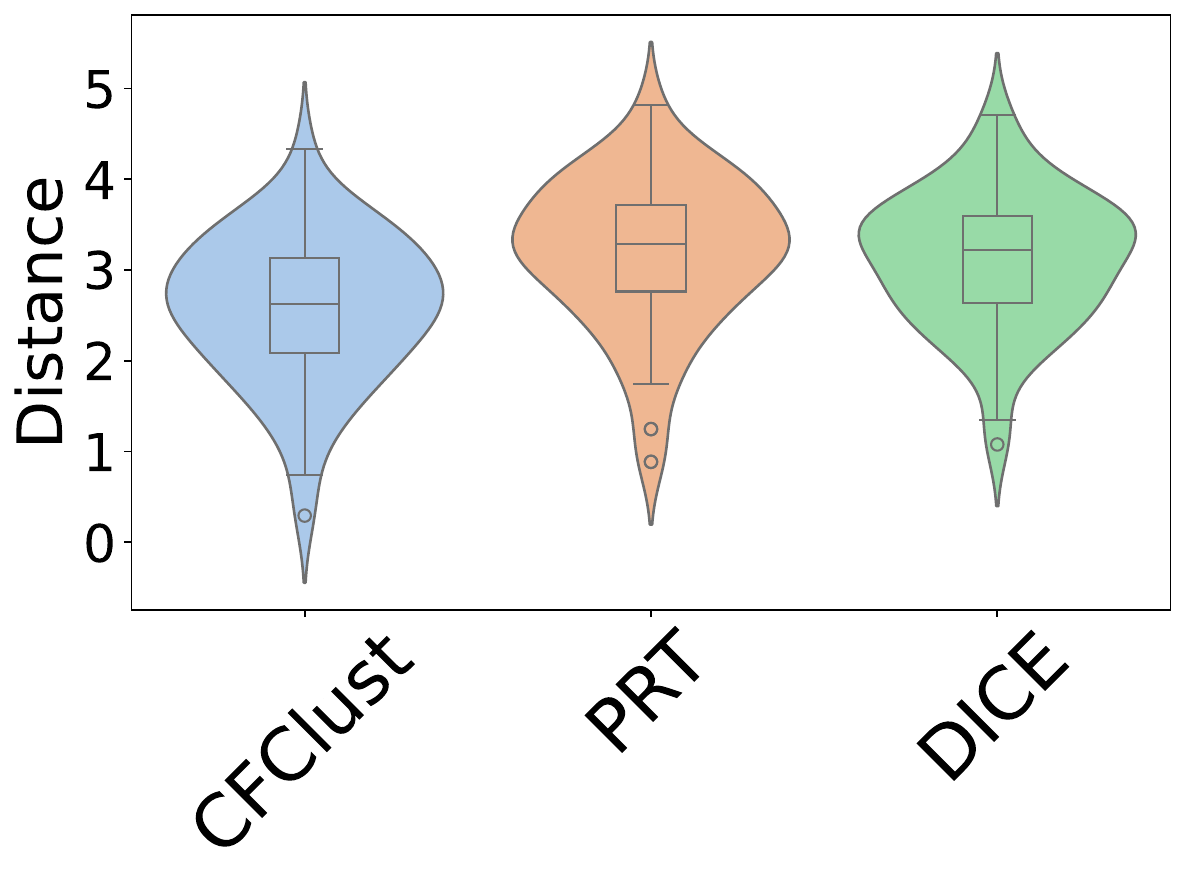} 
                \label{fig:fig:3d-lr-no}
            \end{subfigure}
            \begin{subfigure}[b]{0.3\textwidth}
                \centering
                \includegraphics[width=\textwidth]{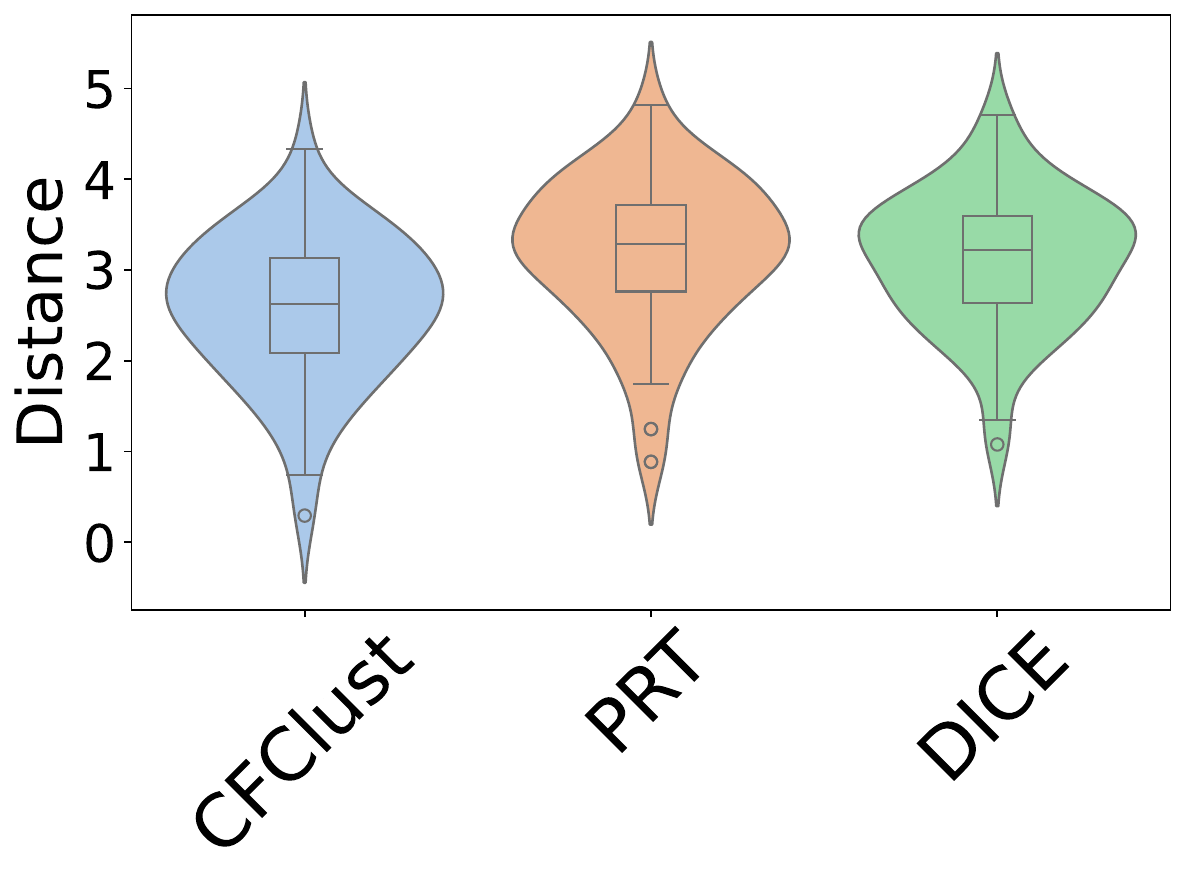} 
                \label{fig:3d-lr-imm2}
            \end{subfigure}
            \begin{subfigure}[b]{0.3\textwidth}
                \centering
                \includegraphics[width=\textwidth]{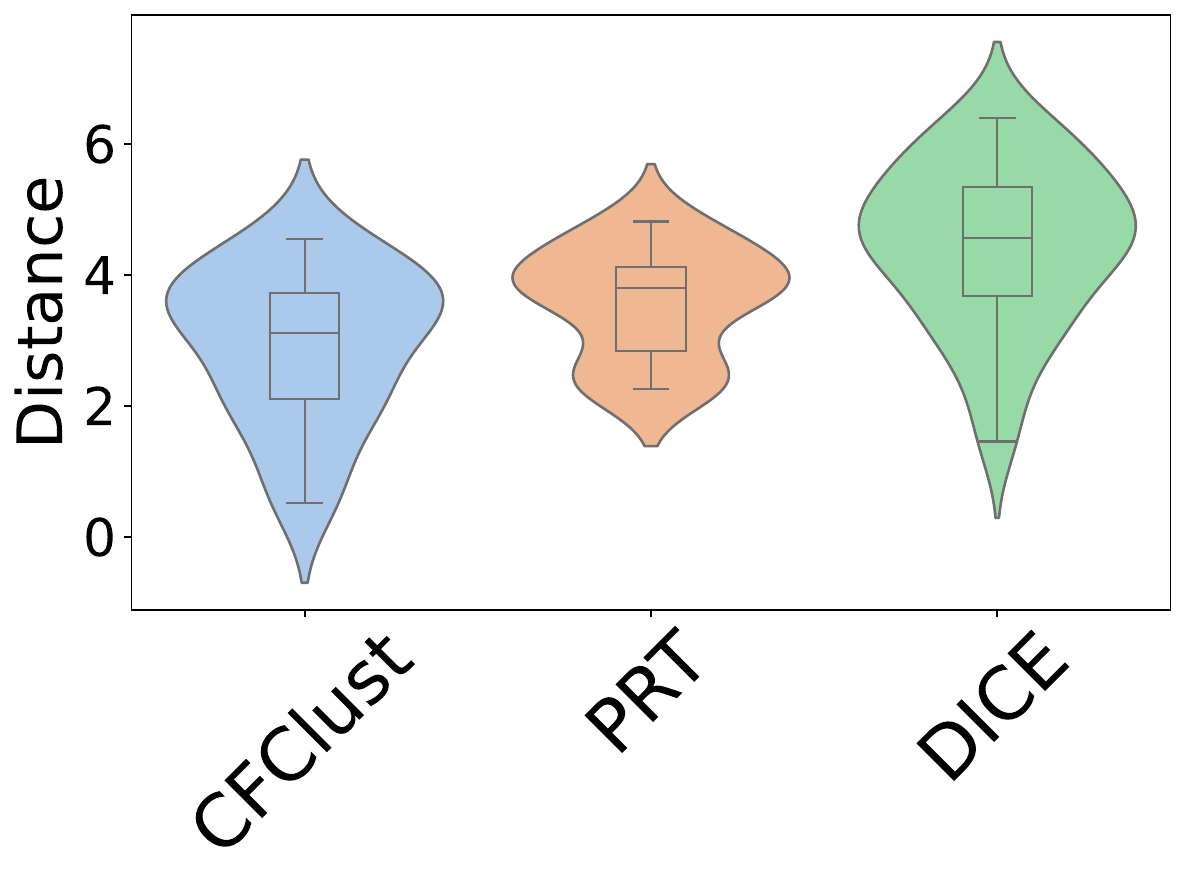} 
                \label{fig:3d-imm13-lr}
            \end{subfigure}
            \hfill
        \end{minipage}
    }
    \caption{$k$-means for {\three}: (left) no immutable features, (middle) one immutable feature,(right) two immutable feature.}
    \label{fig:lr-3d}
\end{figure}

\begin{figure}[h!]
    \centering
    \resizebox{0.9\textwidth}{!}{ 
        \begin{minipage}{\textwidth} 
        \centering
            \begin{subfigure}[b]{0.3\textwidth}
                \centering
                \includegraphics[width=\textwidth]{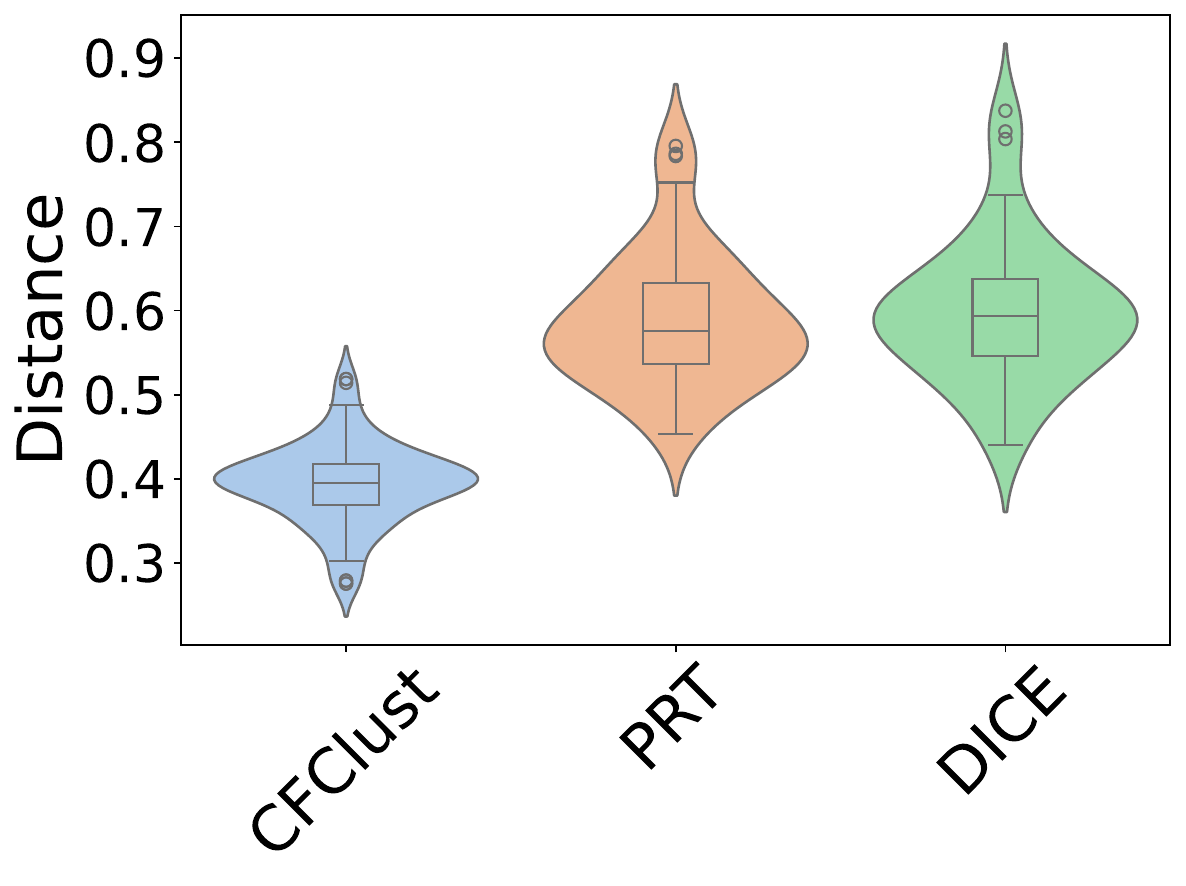} 
                \label{fig:fig:iris-lr-no}
            \end{subfigure}
            \begin{subfigure}[b]{0.3\textwidth}
                \centering
                \includegraphics[width=\textwidth]{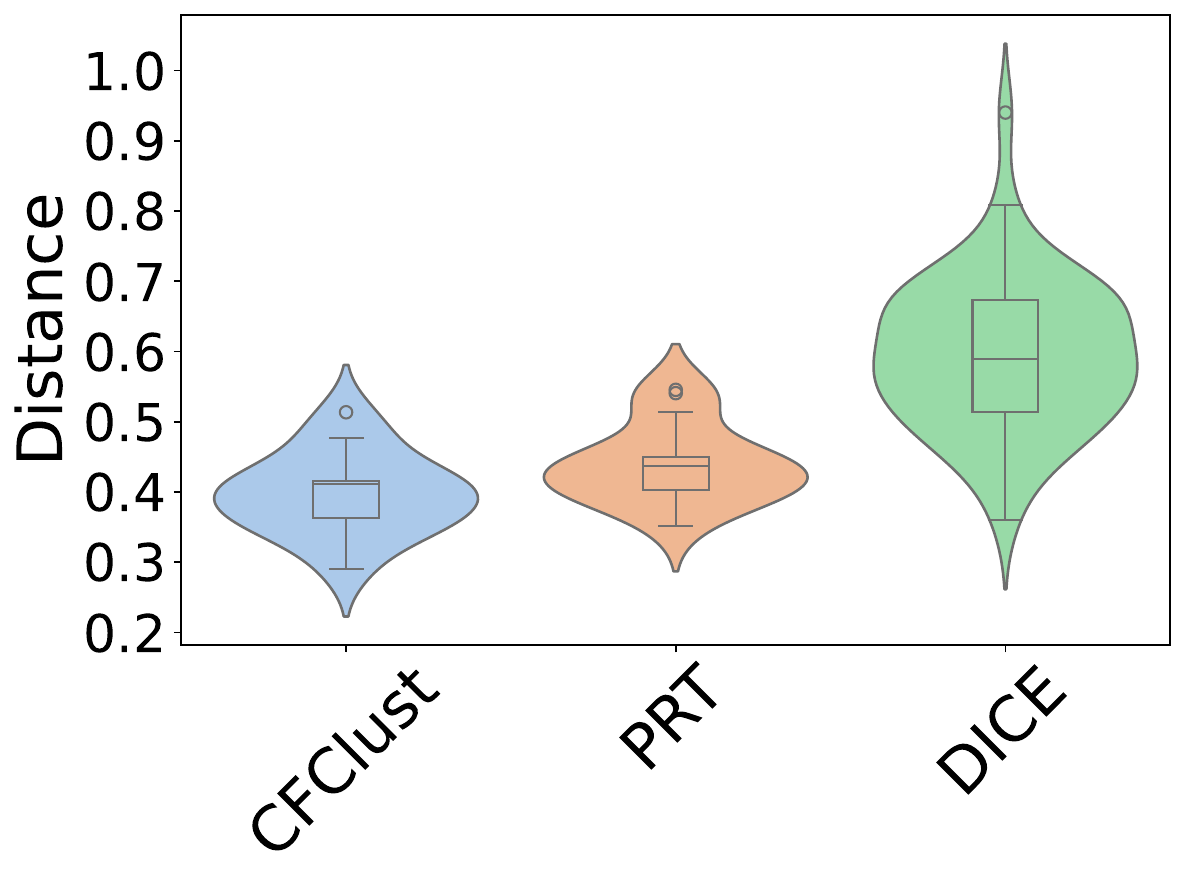} 
                \label{fig:iris-lr-imm1}
            \end{subfigure}
            \begin{subfigure}[b]{0.3\textwidth}
                \centering
                \includegraphics[width=\textwidth]{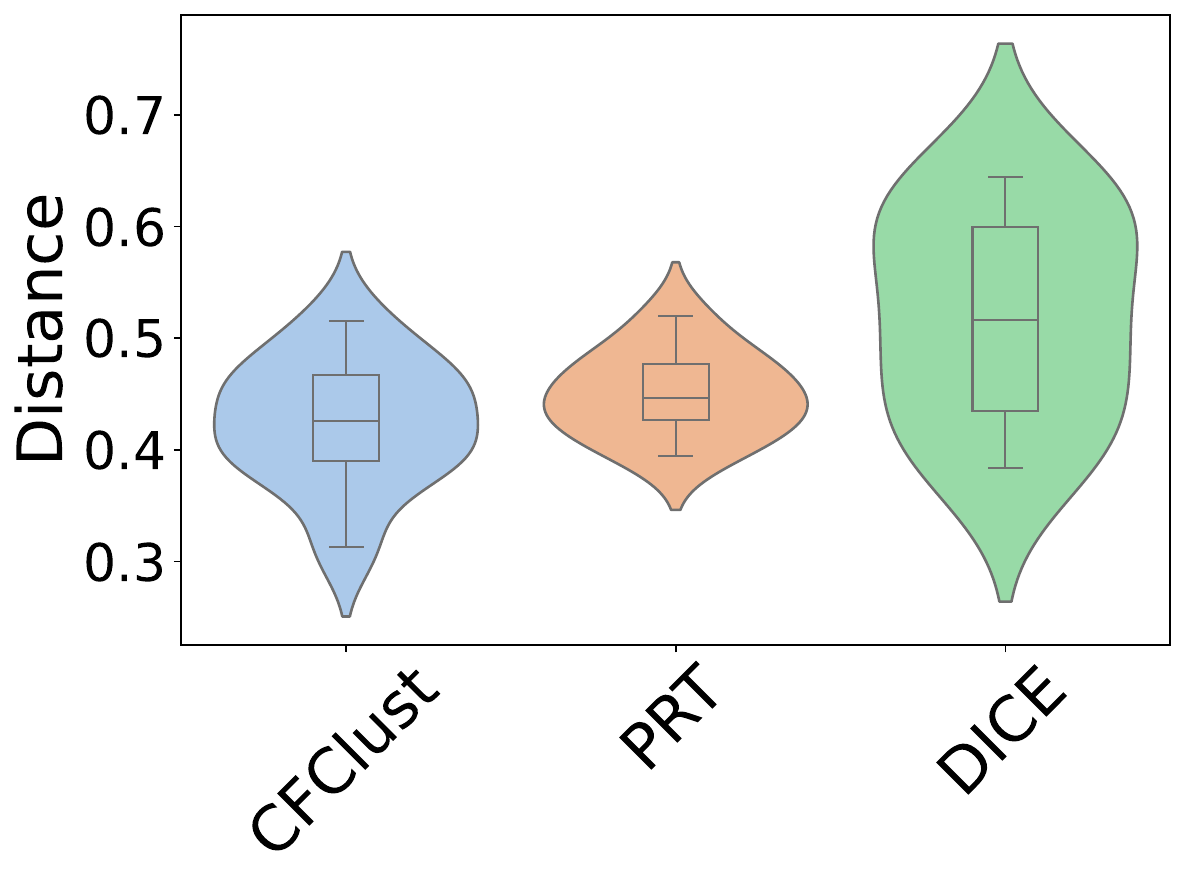} 
                \label{fig:iris-imm12-lr}
            \end{subfigure}
            \hfill
        \end{minipage}
    }
    \caption{$k$-means for {\Iris}: (left) no immutable features, (middle) one immutable feature, (right) two immutable features.}
    \label{fig:lr-iris}
\end{figure}

\begin{figure}[h!]
    \centering
    \resizebox{0.9\textwidth}{!}{ 
        \begin{minipage}{\textwidth} 
        \centering
            \begin{subfigure}[b]{0.3\textwidth}
                \centering
                \includegraphics[width=\textwidth]{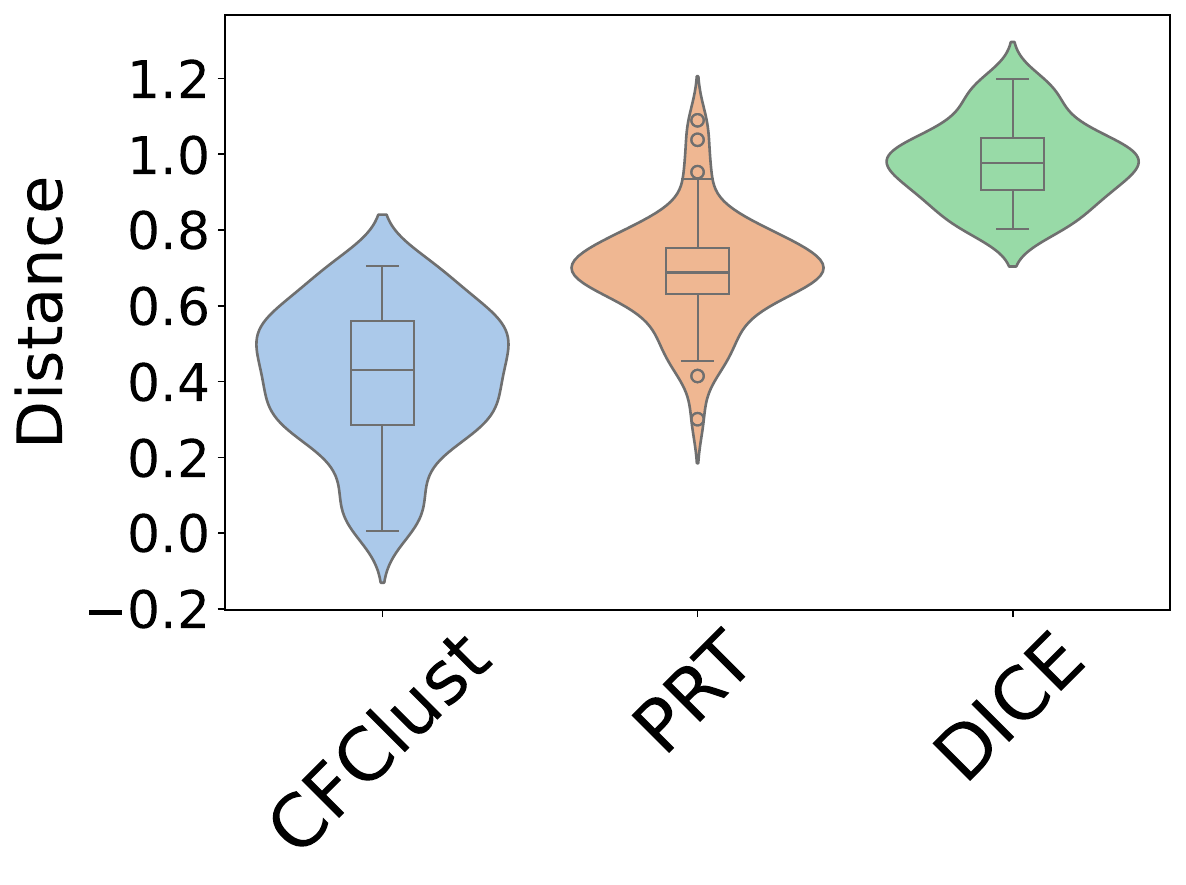} 
                \label{fig:fig:wine-lr-no}
            \end{subfigure}
            \begin{subfigure}[b]{0.3\textwidth}
                \centering
                \includegraphics[width=\textwidth]{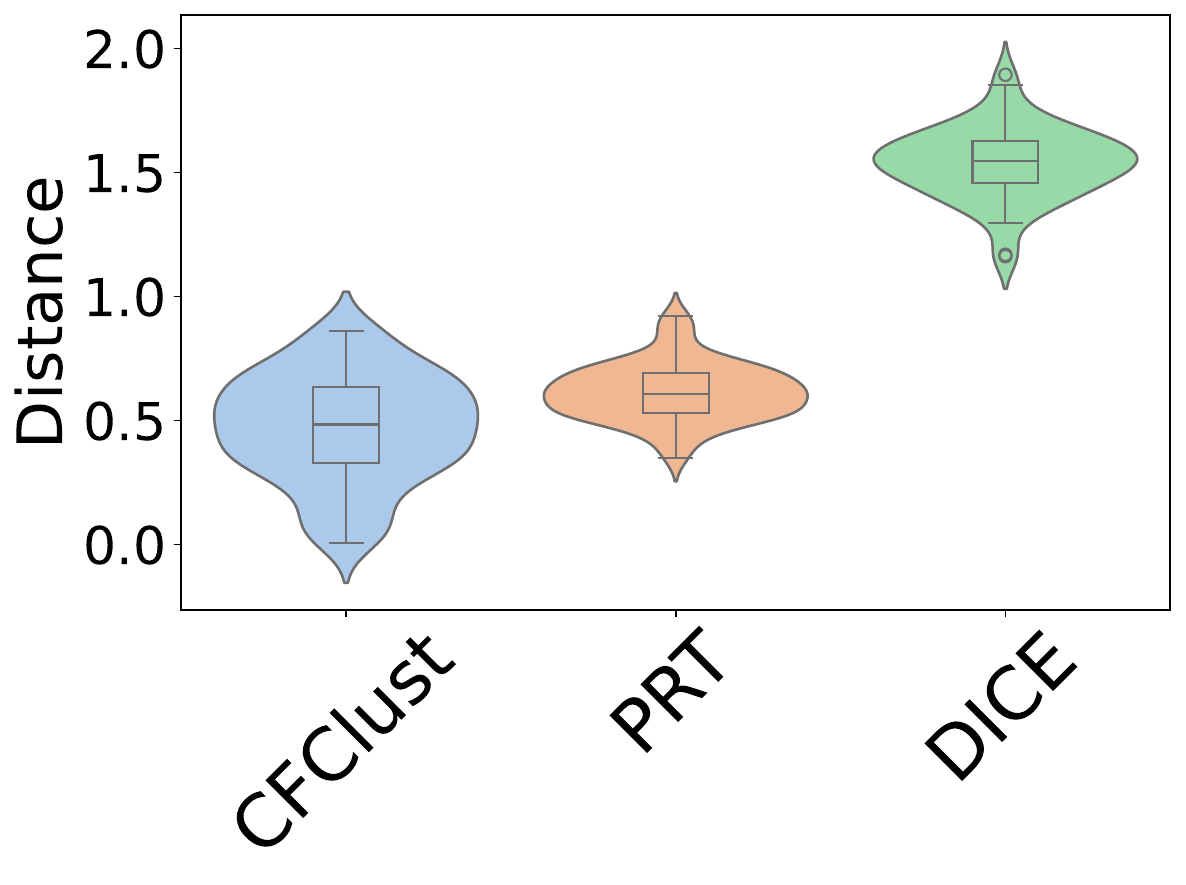} 
                \label{fig:wine-lr-imm}
            \end{subfigure}
            \begin{subfigure}[b]{0.3\textwidth}
                \centering
                \includegraphics[width=\textwidth]{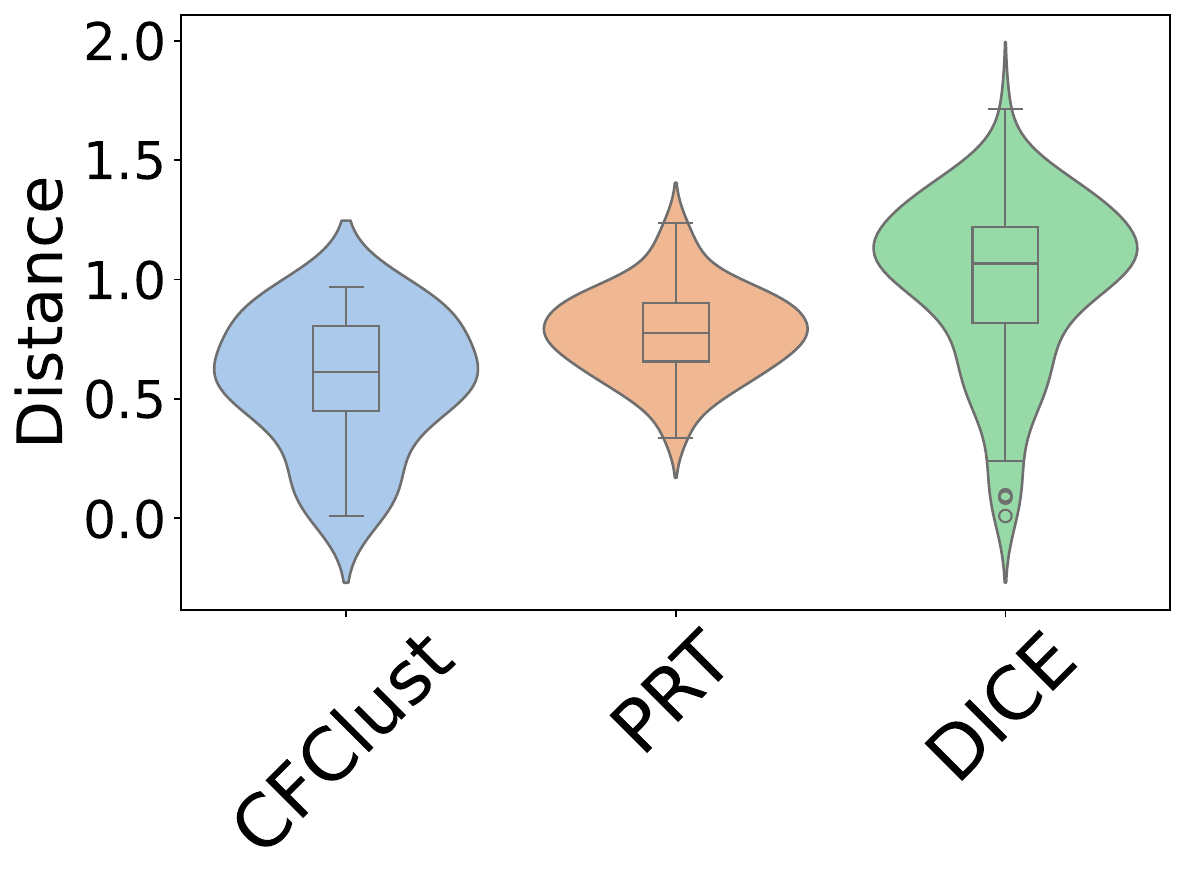} 
                \label{fig:wine-immv2-lr}
            \end{subfigure}
            \hfill
        \end{minipage}
    }
    \caption{$k$-means for {\Wine}: (left) no immutable features, (middle) four immutable feature, (right) seven immutable features.}
    \label{fig:lr-wine}
    \hfill
\end{figure}

\begin{figure}[h!]
    \centering
    \resizebox{0.9\textwidth}{!}{ 
        \begin{minipage}{\textwidth} 
        \centering
            \begin{subfigure}[b]{0.3\textwidth}
                \centering
                \includegraphics[width=\textwidth]{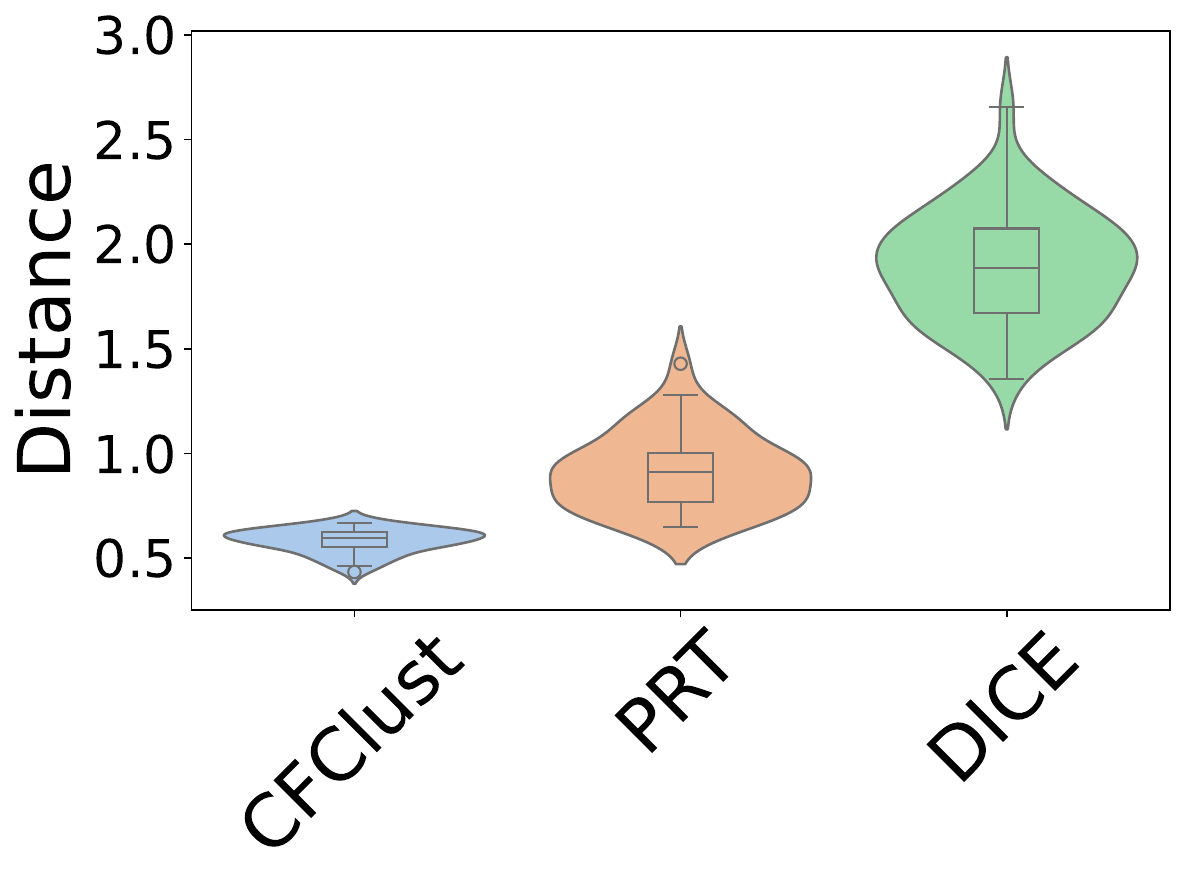} 
                \label{fig:fig:pendigits_lr-no}
            \end{subfigure}
            \begin{subfigure}[b]{0.3\textwidth}
                \centering
                \includegraphics[width=\textwidth]{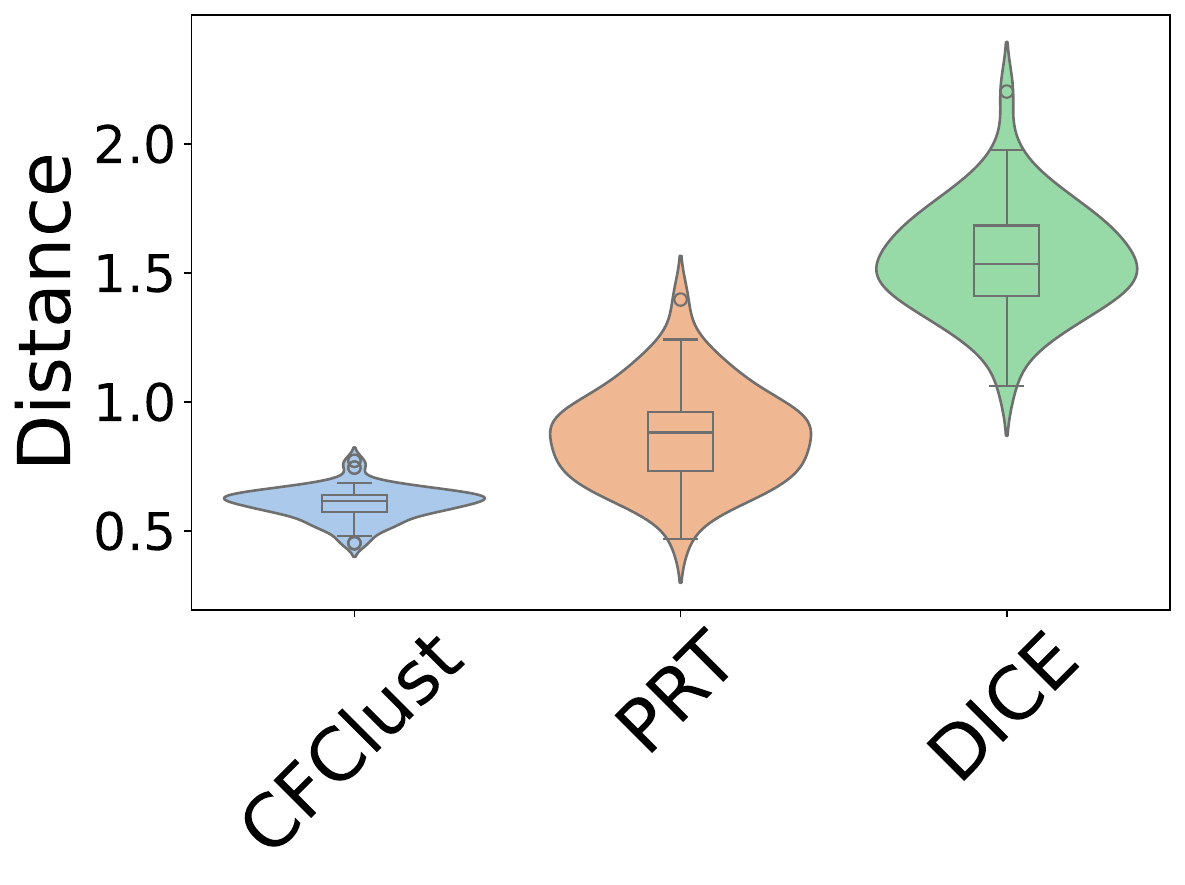} 
                \label{fig:pendigits_imm012_lr}
            \end{subfigure}
            \begin{subfigure}[b]{0.3\textwidth}
                \centering
                \includegraphics[width=\textwidth]{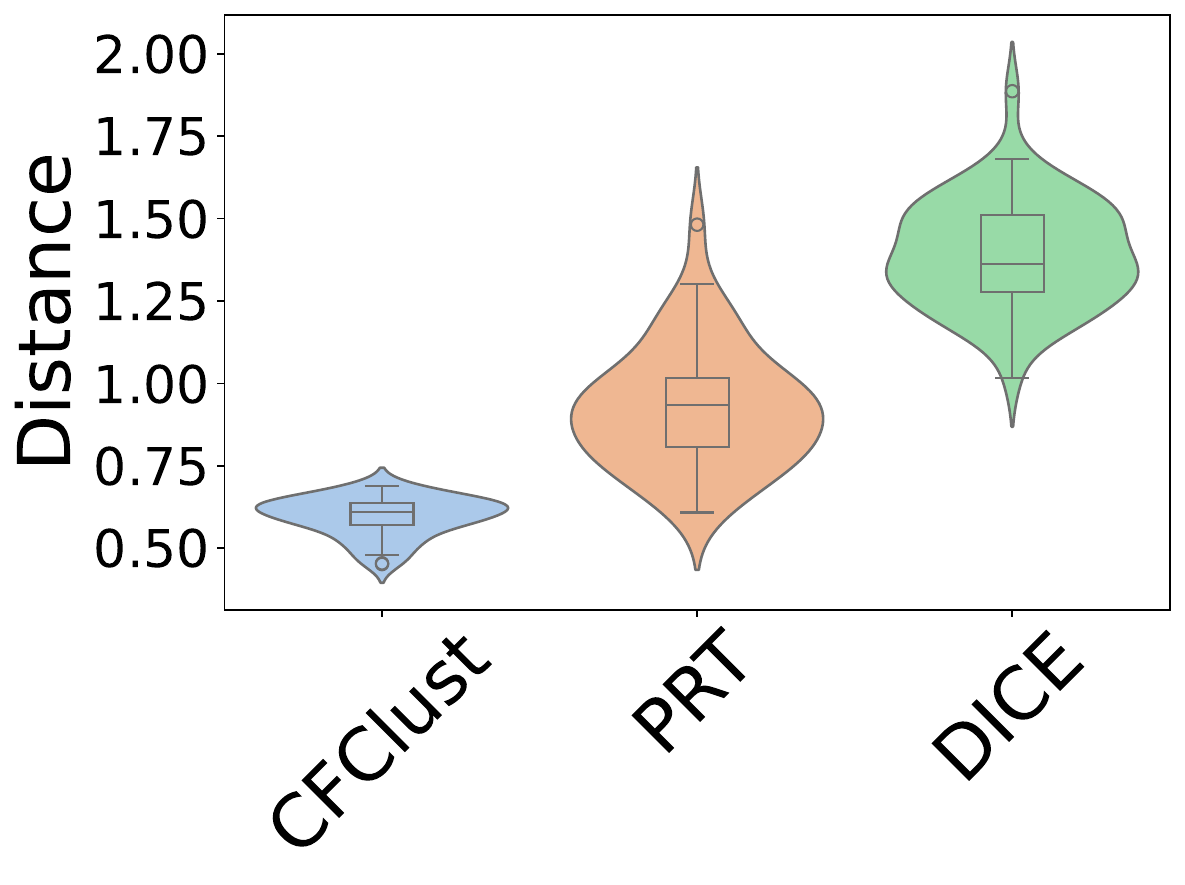} 
                \label{fig:pendigits_imm012345_lr}
            \end{subfigure}
            \hfill
        \end{minipage}
        }
     \caption{$k$-means for {\pendigits}: (left) no immutable features, (middle) three immutable feature, (left) six immutable features.}
    \label{fig:lr-digits}
\end{figure}

We  report the squared Euclidean distance between factual and its generated counterfactual using violin and box plots that provide a detailed representation of the distance distributions.
Fig. \ref{fig:lr-2d} - \ref{fig:lr-digits} depict distance distributions for $k$-means clustering (correspondingly, the $LR$ classifier).
In all cases, our approach generates counterfactuals having the smaller distance.
This is particular evident in datasets with higher number of features.
{\prt} produces counterfactuals with smaller distances than those produced by {\dice}.


\begin{figure}[h!]
    \centering
    \resizebox{0.9\textwidth}{!}{
        \begin{minipage}{\textwidth}
            \centering
            \begin{subfigure}[b]{0.3\textwidth}
                \centering
                \includegraphics[width=\textwidth]{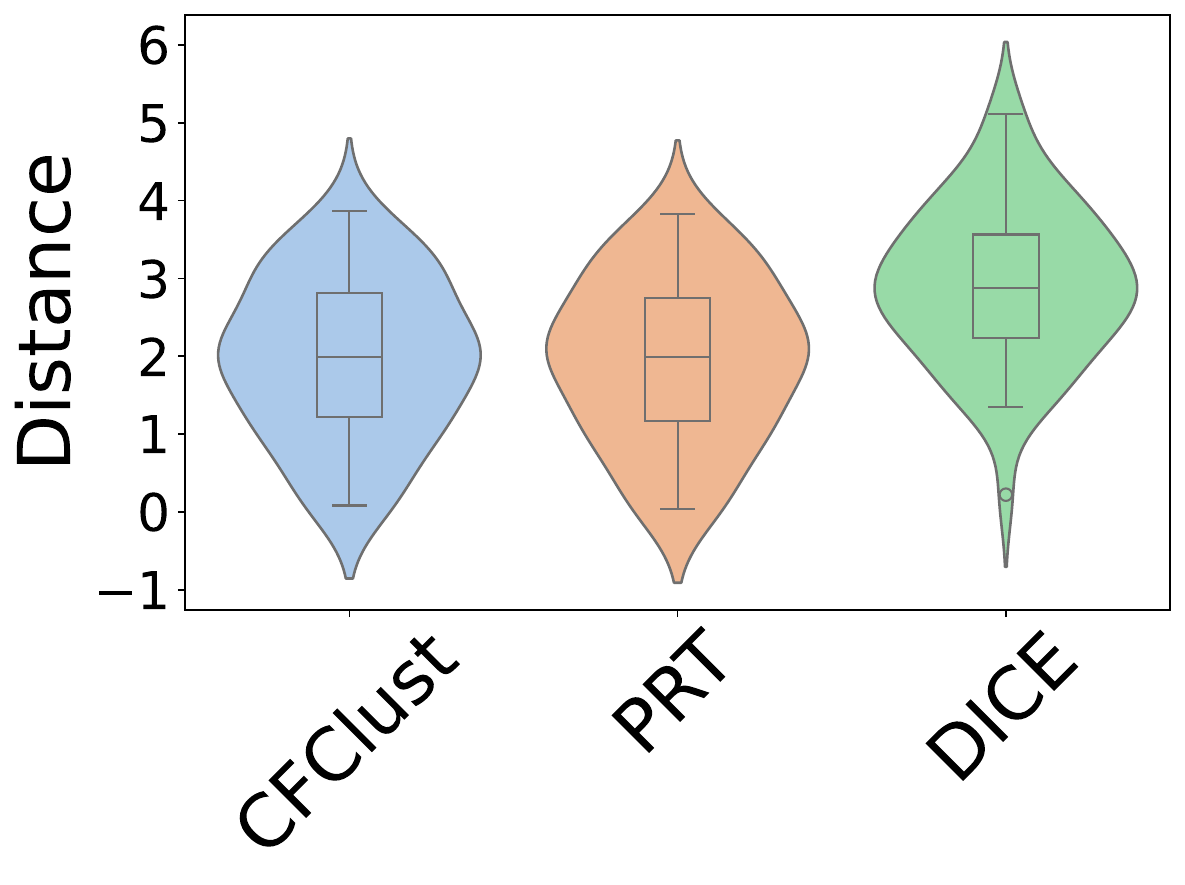} 
                \label{fig:2d-qda}
            \end{subfigure}
            \begin{subfigure}[b]{0.3\textwidth}
                \centering
                \includegraphics[width=\textwidth]{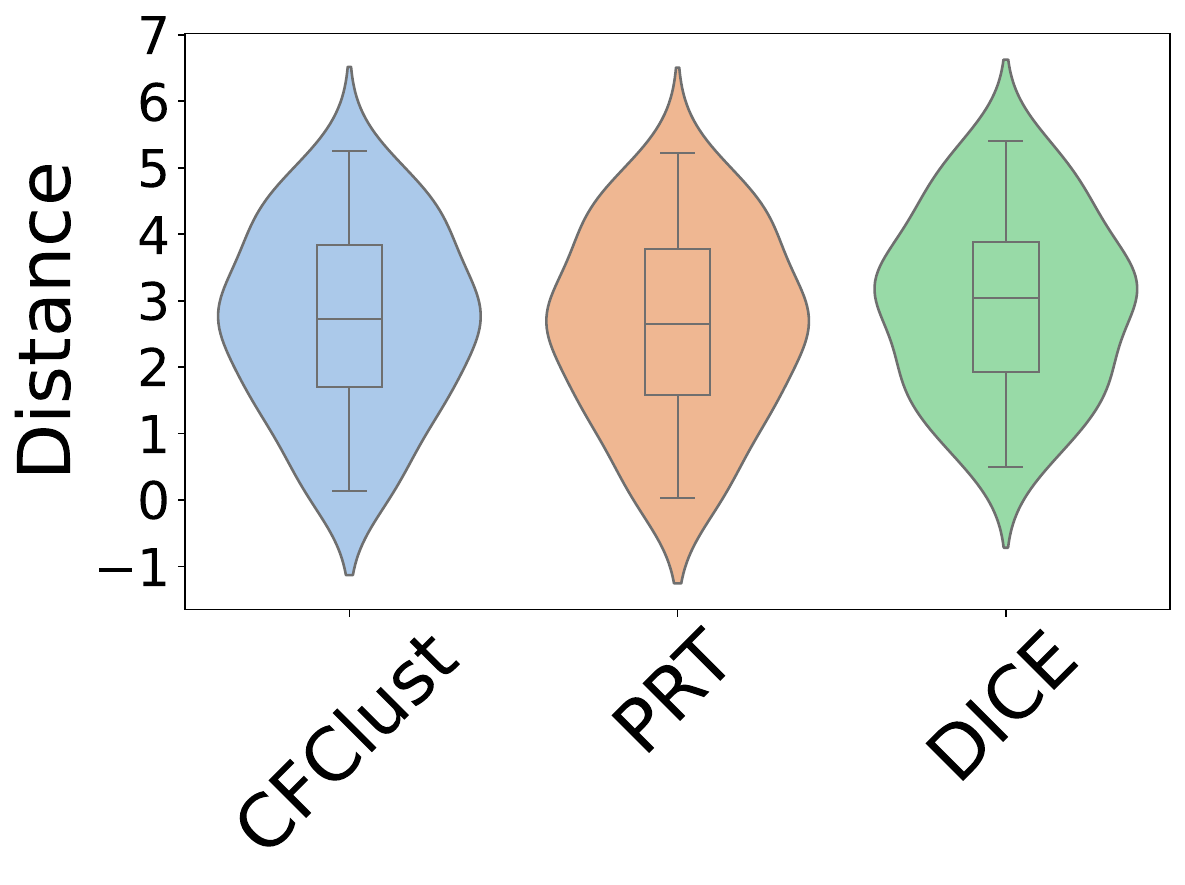} 
                \label{fig:2d-imm1-qda}
            \end{subfigure}
            \begin{subfigure}[b]{0.3\textwidth}
                \centering
                \includegraphics[width=\textwidth]{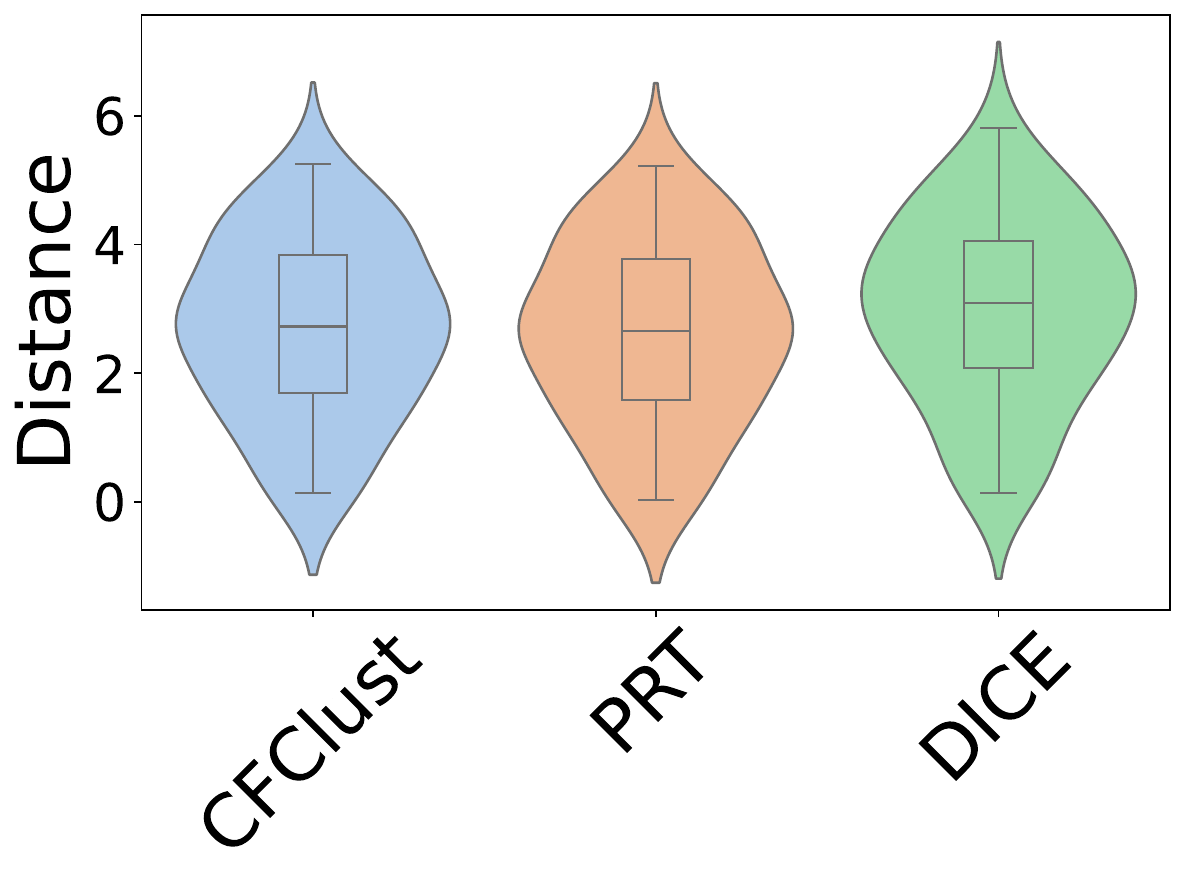} 
                \label{fig:2d-imm2-qda}
            \end{subfigure}
        
        \end{minipage}
        }
        
      \caption{Gaussian clustering for {\two}: (left) no immutable features, (middle)-(right) one immutable feature.}
      \label{fig:qda-2d}
    \hfill
\end{figure}

\begin{figure}[h!]
    \centering
    \resizebox{0.9\textwidth}{!}{
        \begin{minipage}{\textwidth}
            \centering
            \begin{subfigure}[b]{0.3\textwidth}
                \centering
                \includegraphics[width=\textwidth]{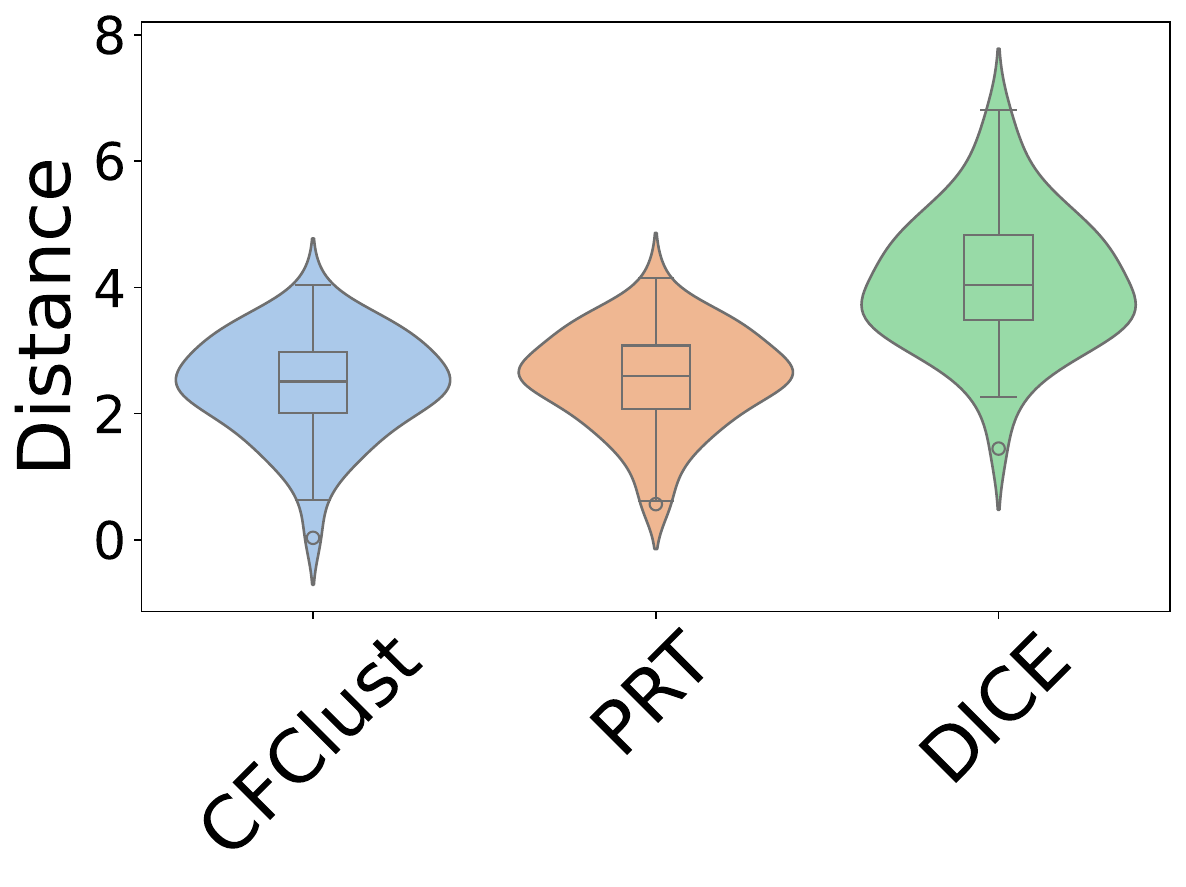} 
                \label{fig:3d-qda}
            \end{subfigure}
            \begin{subfigure}[b]{0.3\textwidth}
                \centering
                \includegraphics[width=\textwidth]{plots/plot_dist_total_3d_imm2.pdf} 
                \label{fig:3d-imm2-qda}
            \end{subfigure}
            \begin{subfigure}[b]{0.3\textwidth}
                \centering
                \includegraphics[width=\textwidth]{plots/plot_dist_total_3d_imm13.pdf} 
                \label{fig:3d-imm13-qda}
            \end{subfigure}
        
        \end{minipage}
        }
      \caption{Gaussian clustering for {\three}: (left) no immutable features, (middle) one immutable feature, (right) two immutable features.}
      \label{fig:qda-3d}
    \hfill
\end{figure}

\begin{figure}[h!]
    \centering
    \resizebox{0.9\textwidth}{!}{ 
        \begin{minipage}{\textwidth}
        \centering
            \begin{subfigure}[b]{0.3\textwidth}
                \centering
                \includegraphics[width=\textwidth]{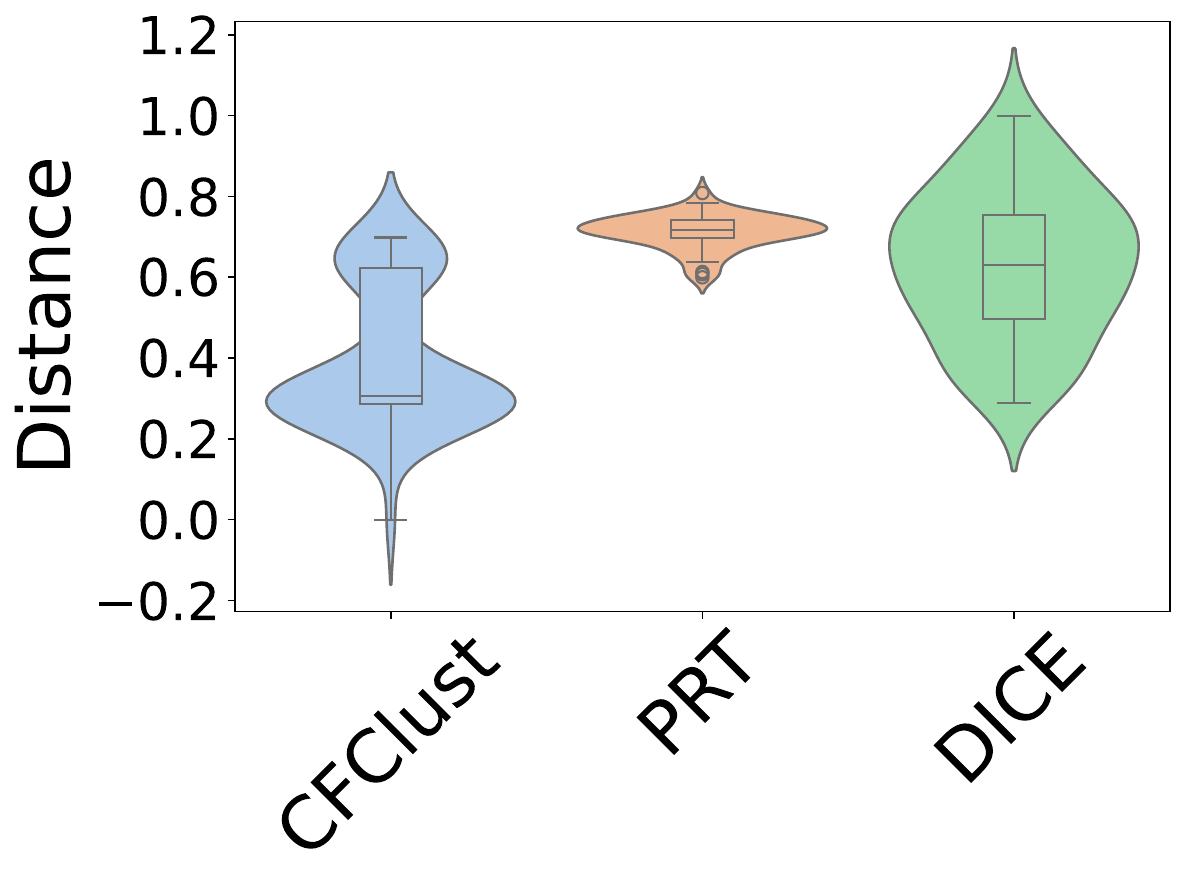} 
                \label{fig:iris-qda}
            \end{subfigure}
            \begin{subfigure}[b]{0.3\textwidth}
                \centering
                \includegraphics[width=\textwidth]{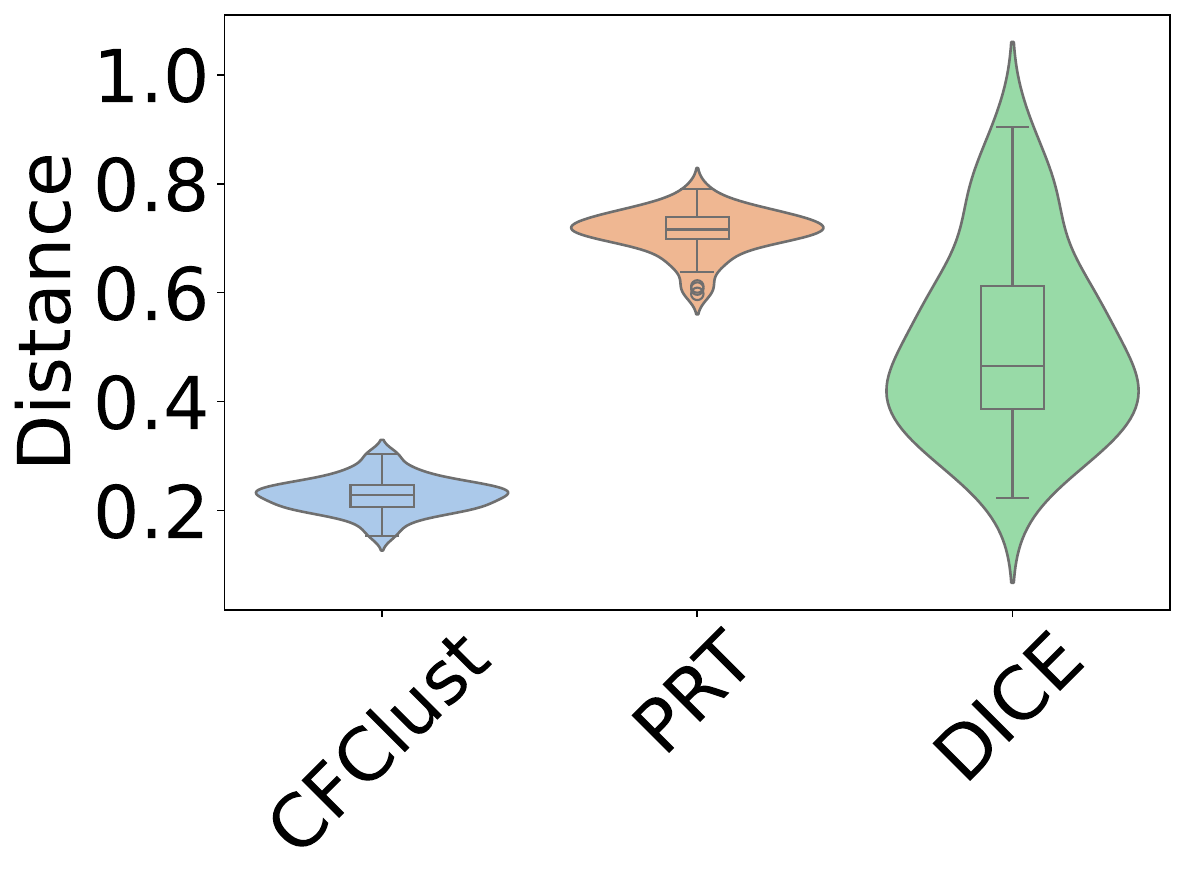} 
                \label{fig:iris-imm1-qda}
            \end{subfigure}
            \begin{subfigure}[b]{0.3\textwidth}
                \centering
                \includegraphics[width=\textwidth]{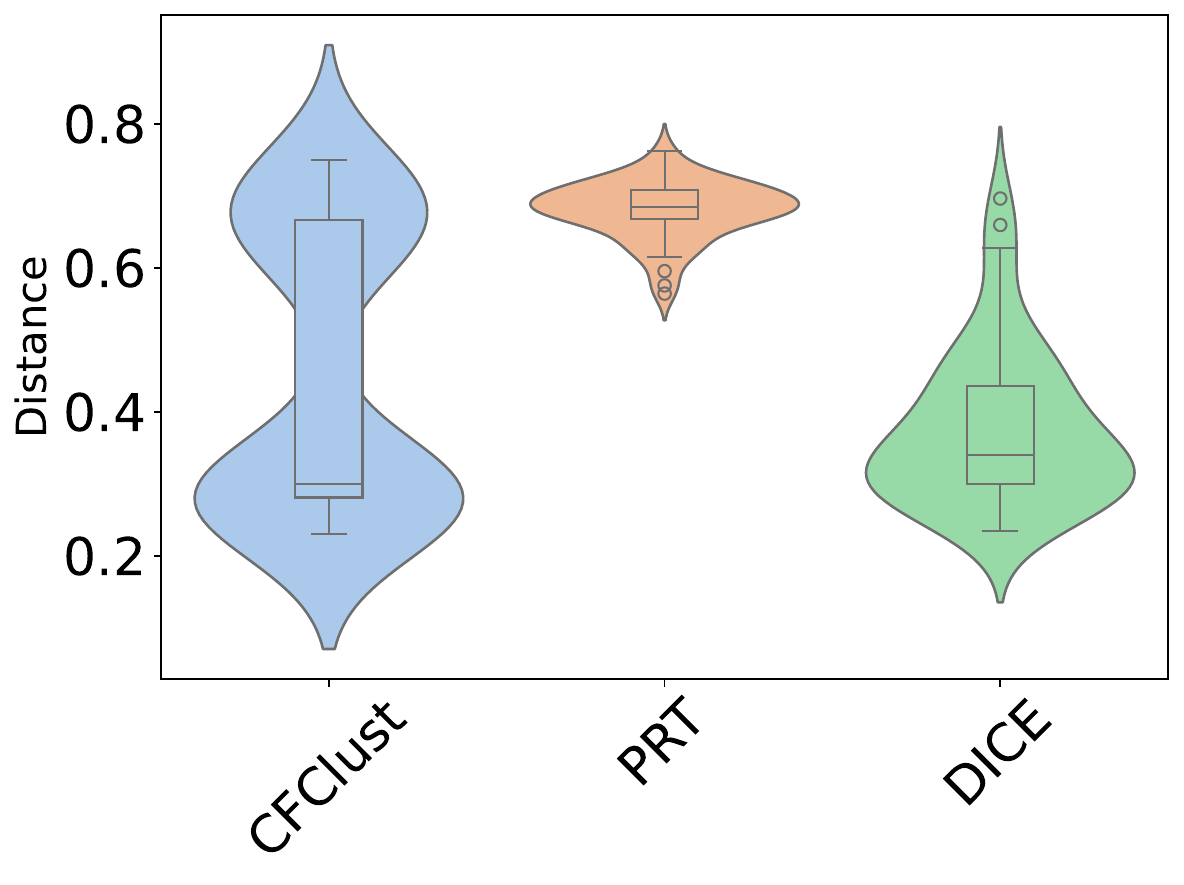} 
                \label{fig:iris-imm12-qda}
            \end{subfigure}
        \end{minipage}
        }
    
    \caption{Gaussian clustering for {\Iris}: (left) no immutable features, (middle) one immutable feature, (right) two immutable features.}
    \label{fig:qda-iris}
\end{figure}

\begin{figure}[h!]
    \centering
    \resizebox{0.9\textwidth}{!}{
        \begin{minipage}{\textwidth}
            \centering
            \begin{subfigure}[b]{0.3\textwidth}
                \centering
                \includegraphics[width=\textwidth]{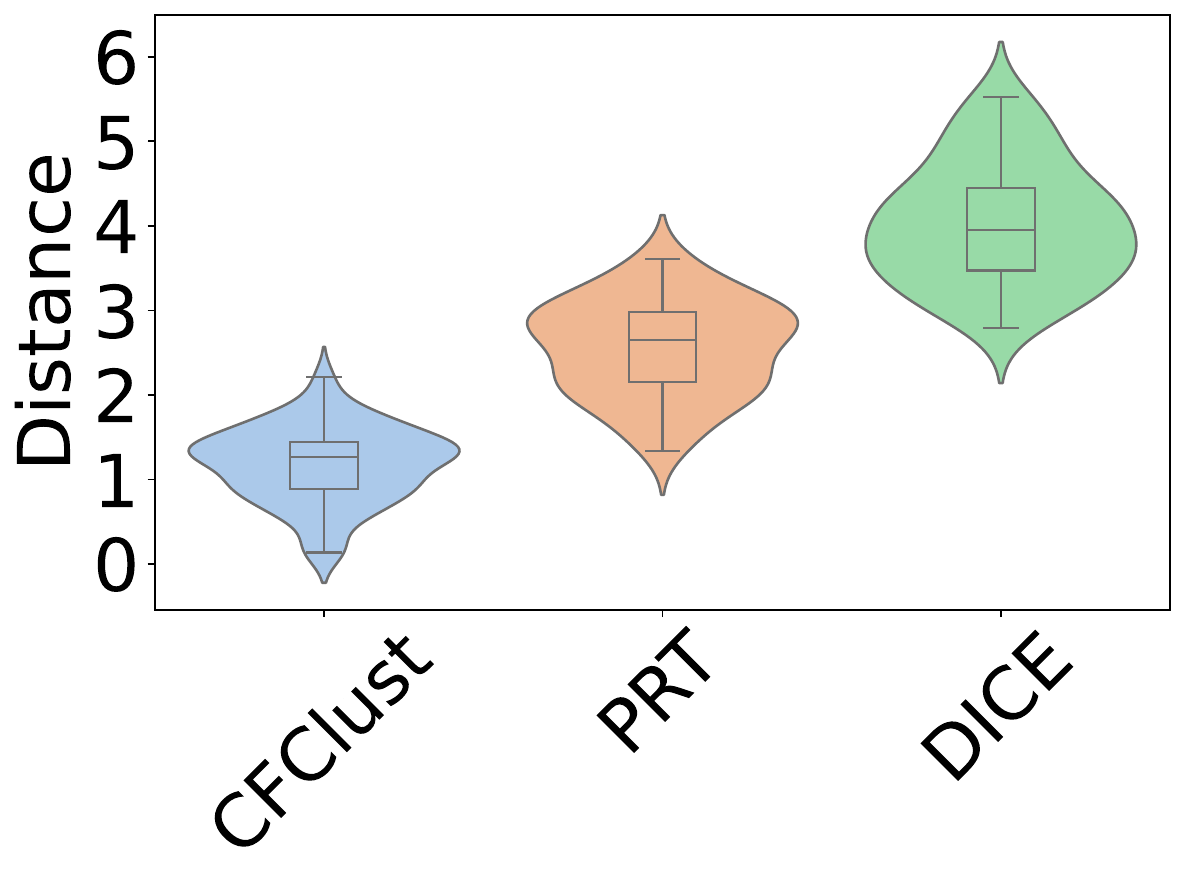} 
                \label{fig:wine-qda}
            \end{subfigure}
            \begin{subfigure}[b]{0.3\textwidth}
                \centering
                \includegraphics[width=\textwidth]{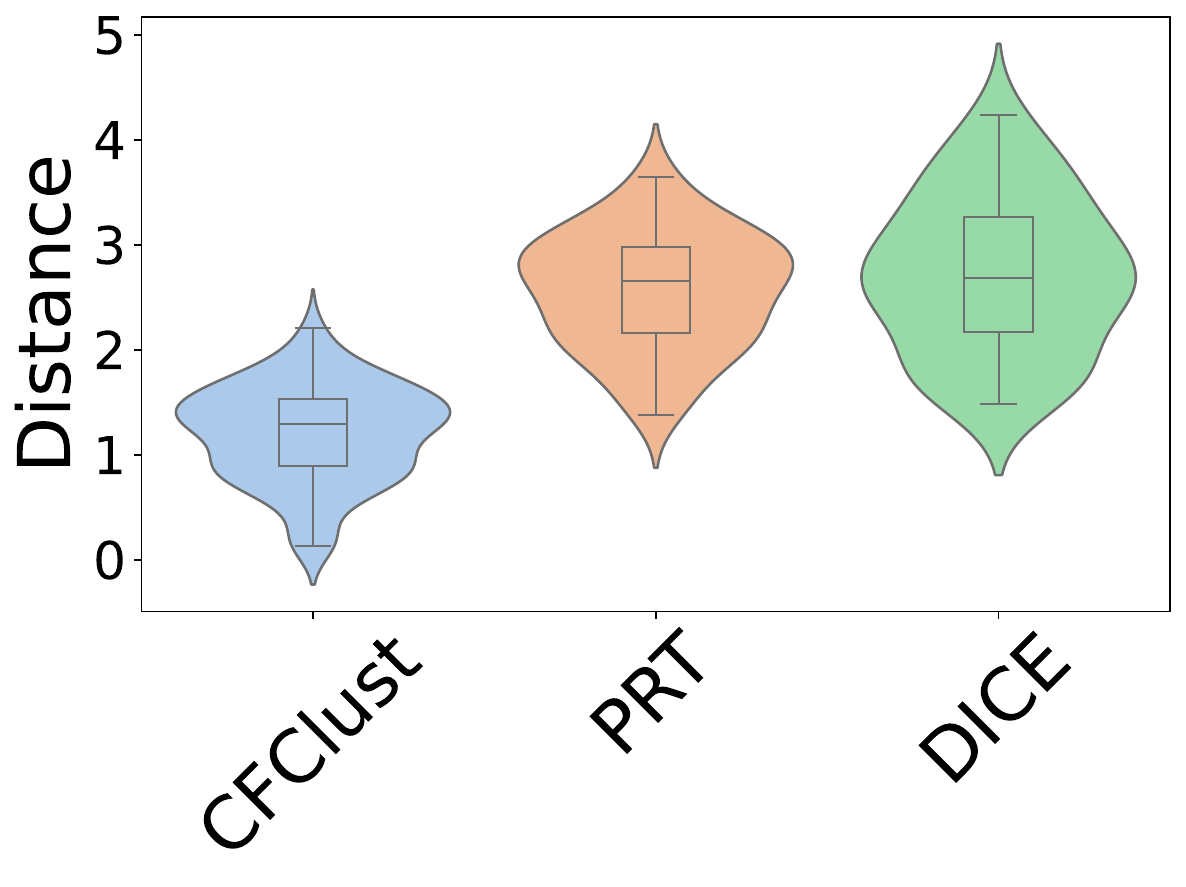} 
                \label{fig:wine-imm-qda}
            \end{subfigure}
            \begin{subfigure}[b]{0.3\textwidth}
                \centering
                \includegraphics[width=\textwidth]{plots/plot_dist_total_wine_imm_v2.pdf} 
                \label{fig:wine-imm_v2-qda}
            \end{subfigure}
        
        \end{minipage}
        }
    \caption{Gaussian clustering for {\Wine}: (left) No immutable features, (middle) four immutable feature, (right) seven immutable features.} \label{fig:qda-wine}
\end{figure}

\begin{figure}[h!]
    \centering
    \resizebox{0.9\textwidth}{!}{
        \begin{minipage}{\textwidth}
            \centering
            \begin{subfigure}[b]{0.3\textwidth}
                \centering
                \includegraphics[width=\textwidth]{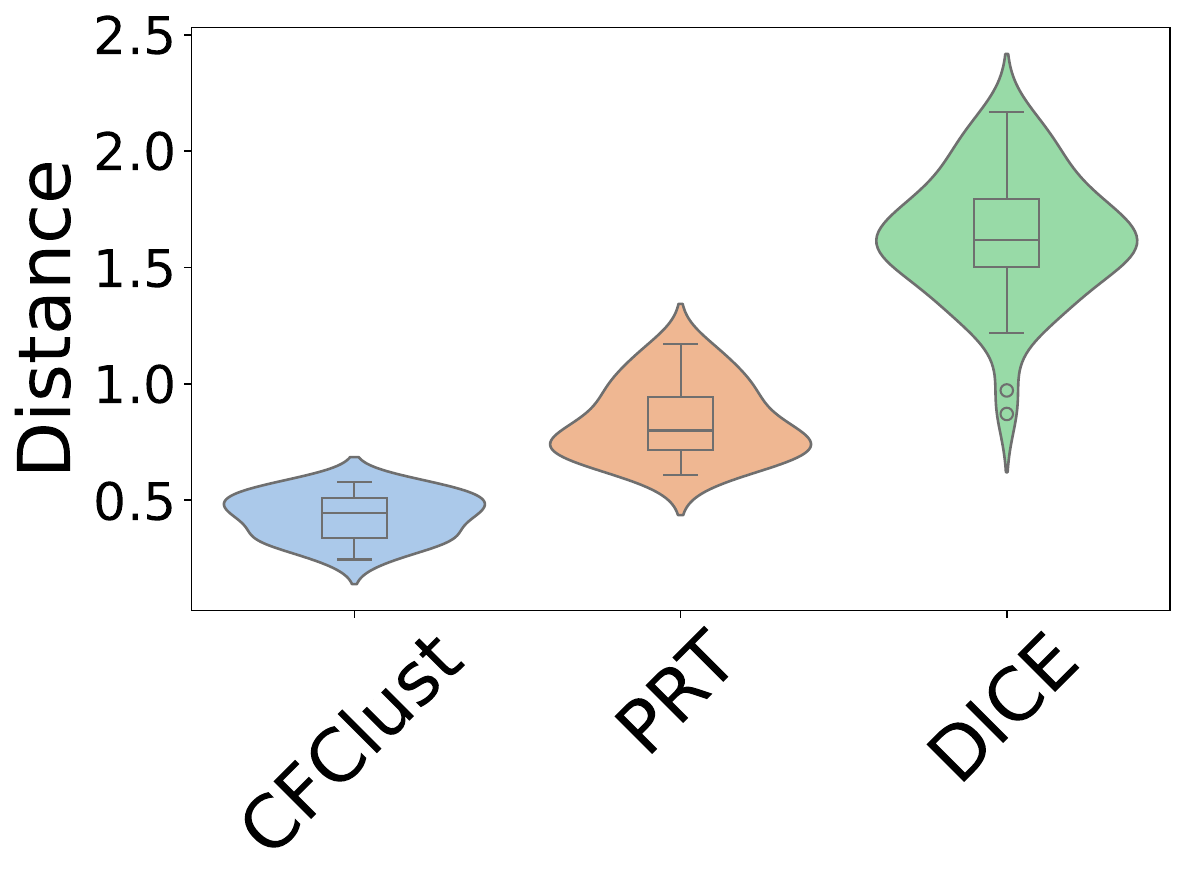} 
                \label{fig:pendigits-qda}
            \end{subfigure}
            \begin{subfigure}[b]{0.3\textwidth}
                \centering
                \includegraphics[width=\textwidth]{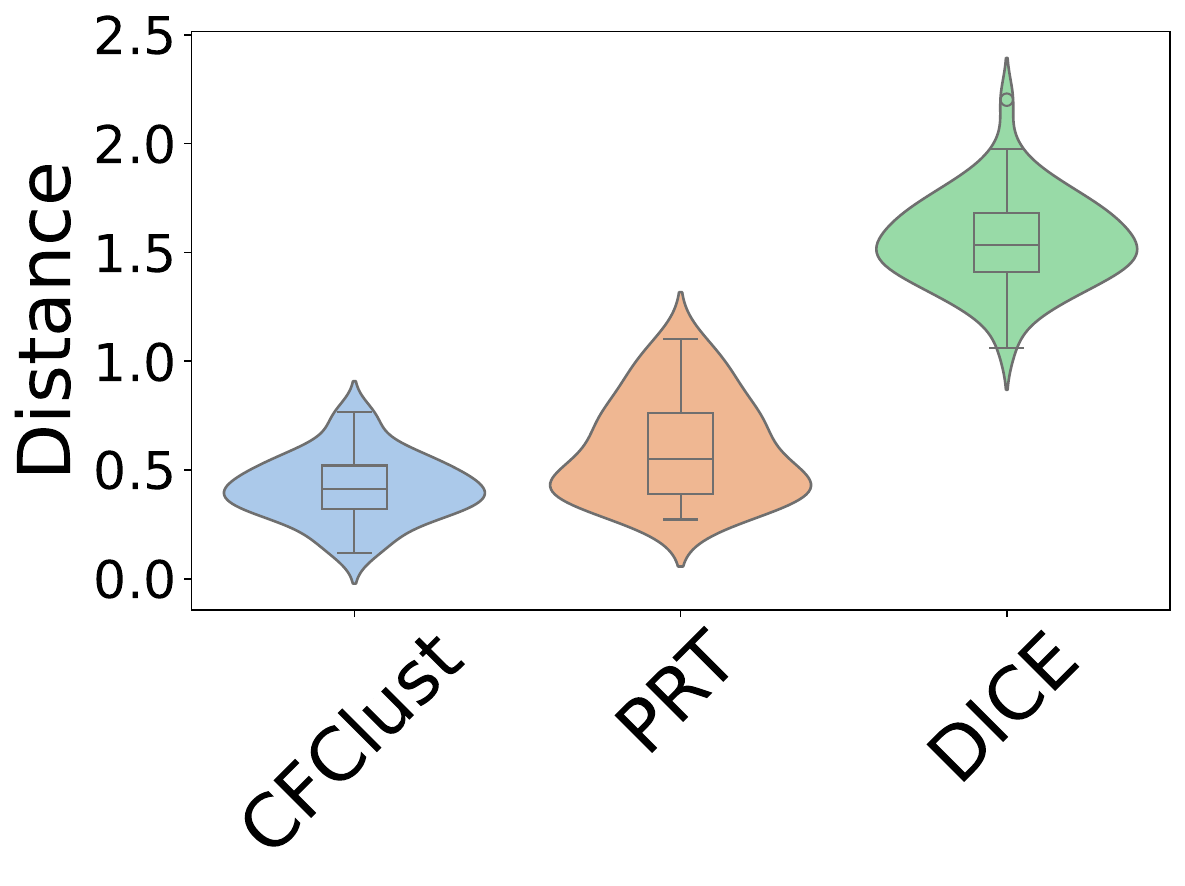} 
                \label{fig:pendigits_imm012_qda}
            \end{subfigure}
            \begin{subfigure}[b]{0.3\textwidth}
                \centering
                \includegraphics[width=\textwidth]{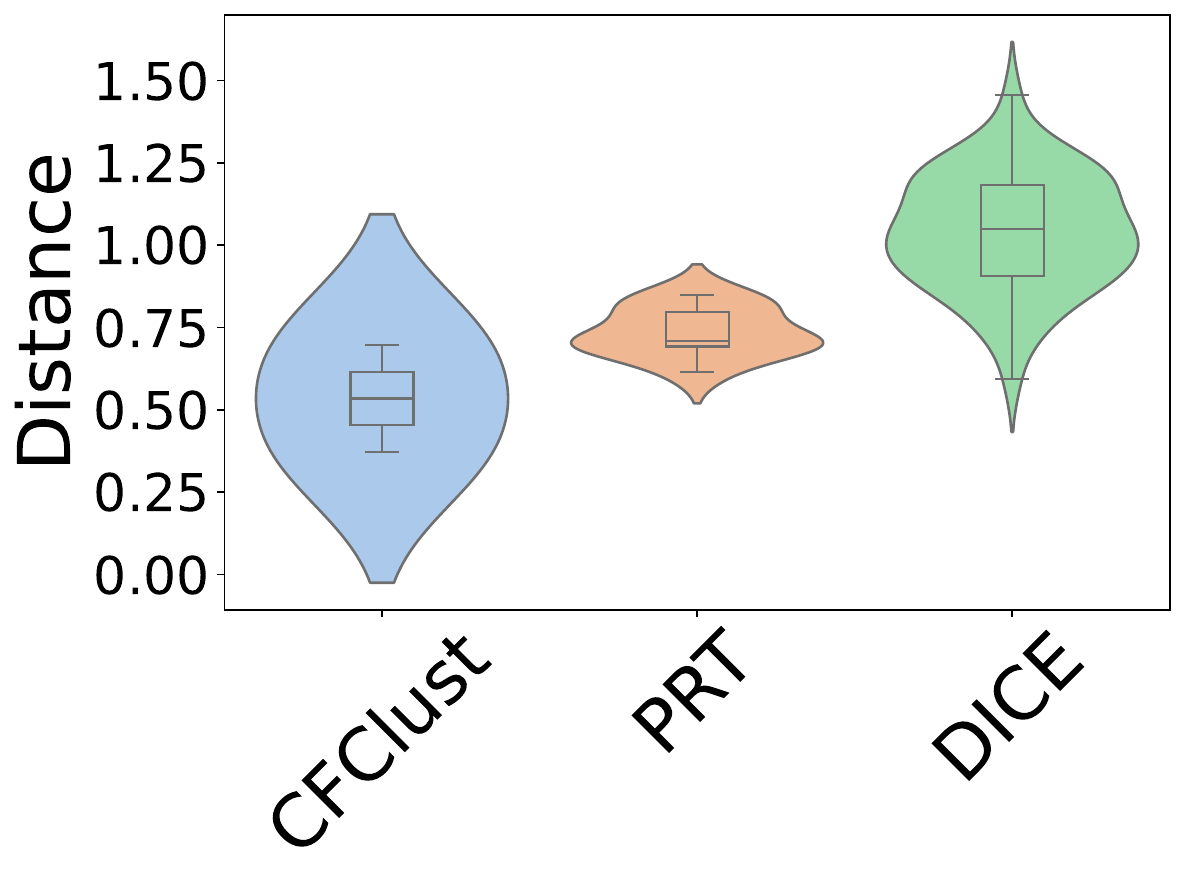} 
                \label{fig:pendigits_imm012345_qda}
            \end{subfigure}
        \end{minipage}
        }
    \caption{Gaussian for {\pendigits}: (left) no immutable features, (middle) three immutable feature, (right) six immutable features.}
      \label{fig:qda-digits}
\end{figure}

Fig. \ref{fig:qda-2d} - \ref{fig:qda-digits} depict counterfactual distance distributions for Gaussian clustering (correspondingly, the $QDA$ classifier). Again, it is evident that our approach provides counterfactuals of smaller distance.

In Table \ref{tab:t-test-lr-qda} we provide results from t-test comparison of {\cfclust} to each other method. The statistical significance level ($\alpha$) is set to 0.05. The symbol `+' in the table indicates that {\cfclust} is found superior to the compared method, while the symbol `=' indicates no statistically significant difference. As the table indicates, in all but a few equal cases corresponding to simpler low dimensional problems, 
the superiority of {\cfclust} is statistically significant. As dimensionality increases, optimization becomes more challenging, and our method's superiority is more clear. 

\begin{table}[htbp]
\scriptsize
\centering
\caption{Statistical significance comparison of {\cfclust} to other methods using t-test.}
\label{tab:t-test-lr-qda}
\begin{tabular}{lcccc} 
\toprule
Method & {\prt}(LR) & {\dice}(LR) & {\prt}(QDA) & {\dice}(QDA) \\ 
\midrule
2D      & = & = & + & +  \\ 
2D Imm. Feat. $1$ & = & = & = & = \\
2D Imm. Feat. $2$ & = & = & = & = \\
3D & + &  +& + &  +\\
3D Imm. Feat. $2$ &+ & = & + &  +\\
3D Imm. Feat. $1,3$ &+ & + & = &  =\\
Iris & +  &  +& + &  +\\
Iris Imm. Feat. $1$ & = & +& + &  +\\
Iris Imm. Feat. $1,2$ & = & +& + &  +\\
Wine & + & + & + &  +\\
\makecell[l]{Wine Imm. Feat. \\  $1,2,5,10$} & + & +& + &  + \\
\makecell[l]{Wine Imm. Feat. \\  $1,2,5,10-13$} & + & =& = & +  \\
Pendigits & + & + & + &  +\\
Pendigits & + & + & + &  +\\
\makecell[l]{Pendigits Imm. Feat. \\  $0,1,2$} & +  & + & + &  + \\
\makecell[l]{Pendigits Imm. Feat. \\  $0,1,2,3,4,5$} & + & + & + & +  \\
\bottomrule
\end{tabular}
\end{table}

Finally, we report results regarding execution time in Table \ref{tab:time-lr-qda}. 
While our method requires less than 0.001 seconds average time for generating a counterfactual, the other methods are significantly slower. It should also be noted that {\dice} is much faster than {\prt}.


\begin{table}[htbp]
\scriptsize
\centering
\caption{Average time (seconds) for generating a single counterfactual. In all cases, the {\cfclust} execution time is below 0.001 seconds.}
\label{tab:time-lr-qda}
\begin{tabular}{lrrrr} 
\toprule
Method & PRT(LR) & DICE(LR) & PRT(QDA) & DICE(QDA)  \\ 
\midrule
2D & 149.43 & 0.04 & 165.32 & 0.04 \\ 
2D Imm. Feat. $1$ & 735.75 & 0.05 & 218.66 &0.04 \\
2D Imm. Feat. $2$ & 183.39 & 0.05 & 218.66&0.07 \\ 
3D & 104.45 &  0.06 & 128.99 & 0.06 \\
3D Imm. Feat. $2$ & 499.81 & 0.07 & 259.19& 0.06 \\
3D Imm. Feat. $1,3$ & 540.49  & 0.06 & 282.98 & 0.13 \\
Iris & 149.43  &  0.05 & 80.47 & 0.05 \\
Iris Imm. Feat. $1$ & 57.64 & 0.05 & 55.54 & 0.06\\
Iris Imm. Feat. $1,2$ & 82.08 & 0.08 & 121.83 & 0.07\\
Wine & 159.97 & 0.11 & 78.49 & 0.11 \\
\makecell[l]{Wine Imm. Feat. \\  $1,2,5,10$} & 263.57 & 0.12 & 79.83 & 0.12  \\
\makecell[l]{Wine Imm. Feat. \\  $1,2,5,10-13$} & 84.61 & 0.04  & 93.69  & 0.08  \\

Pendigits & 97.43 & 0.12 & 78.50 & 0.083 \\
\makecell[l]{Pendigits Imm. Feat. \\  $0,1,2$} & 129.27 & 0.38& 138.16 & 0.37  \\
\makecell[l]{Pendigits Imm. Feat. \\  $0,1,2,3,4,5$} &  143.53 & 0.71 & 132.62 & 0.21 \\
\bottomrule
\end{tabular}
\end{table}

In summary, it is evident that the proposed {\cfclust} approach provides superior solutions at negligible computation time without requiring any parameter tuning.

\section{Conclusions}
\label{sec:conclusions}
In this work we have considered the use of counterfactuals to explain \emph{clustering} solutions. 
At first we have presented a general definition for counterfactuals for clustering assuming that each cluster is modeled using a probability density. Then we considered the counterfactual generation problem for $k$-means and Gaussian clustering assuming squared Euclidean distance among the factual and the counterfactual. Our method takes as input the factual instance $y$, the target cluster, the cluster model, the feature actionability mask $M$ and the plausibility factor $\epsilon$ and generates the counterfactual $z$ through a simple computational procedure.

More specifically, in the $k$-means clustering case, analytical mathematical formulas are presented for computing the optimal solution taking also into account plausibility and actionability constraints. In the Gaussian clustering case (assuming full, diagonal or spherical covariances) our method requires the numerical solution of a nonlinear equation with a single parameter only. 

A distinct advantage of our method is that it uses analytical or non-iterative solutions for generating counterfactuals. This is due to the constrained optimization problem we have defined that minimizes the Euclidean distance under linear or quadratic equality constraints. 
Current counterfactual generation methods for classifiers employ iterative methods (either very general or more specific) to solve the corresponding constrained optimization problem providing solutions that may be suboptimal and require parameter tuning (regularization parameters, number of iterations etc). In our experimental study, we conducted comparisons with an indirect approach where we treat the clustering problem as a classification one (either logistic regression or quadratic discriminant analysis) and employ popular tools developed for classification problems. The experimental comparison reveals the optimality, effectiveness and convenience of our methodology.  

Future work could focus on extending the method to account for other distance measures such as the $L_1$ norm. In this case direct solutions cannot be obtained and iterative search methods should be employed. It would also be interesting to investigate other cluster models such as kernel-based clustering and adapt the method to generate multiple diverse counterfactuals for a given factual.
Since our method provides fast counterfactual computation, it can be used to efficiently generate counterfactuals for a large number of factuals. Another research direction could focus on expoiting those counterfactuals to quantify significant properties of a clustering solution, such as cluster separability and fairness. 

\section*{Acknowledgment} This research project is implemented in the framework of H.F.R.I. call ``Basic research Financing (Horizontal support of all Sciences)'' under the National Recovery and Resilience Plan ``Greece 2.0'' funded by the European Union - NextGenerationEU (H.F.R.I. ProjectNumber: 15940).

\bibliographystyle{unsrt}  
\bibliography{bibliography}

\begin{thebibliography}{10}

\bibitem{ribeiro2016should}
Marco~Tulio Ribeiro, Sameer Singh, and Carlos Guestrin.
\newblock " why should i trust you?" explaining the predictions of any classifier.
\newblock In {\em Proceedings of the 22nd ACM SIGKDD international conference on knowledge discovery and data mining}, pages 1135--1144, 2016.

\bibitem{wachter2017counterfactual}
Sandra Wachter, Brent Mittelstadt, and Chris Russell.
\newblock Counterfactual explanations without opening the black box: Automated decisions and the gdpr.
\newblock {\em Harv. JL \& Tech.}, 31:841, 2017.

\bibitem{adebayo2018sanity}
Julius Adebayo, Justin Gilmer, Michael Muelly, Ian Goodfellow, Moritz Hardt, and Been Kim.
\newblock Sanity checks for saliency maps.
\newblock {\em Advances in neural information processing systems}, 31, 2018.

\bibitem{guidotti2024counterfactual}
Riccardo Guidotti.
\newblock Counterfactual explanations and how to find them: literature review and benchmarking.
\newblock {\em Data Mining and Knowledge Discovery}, 38(5):2770--2824, 2024.

\bibitem{verma2024counterfactual}
Sahil Verma, Varich Boonsanong, Minh Hoang, Keegan Hines, John Dickerson, and Chirag Shah.
\newblock Counterfactual explanations and algorithmic recourses for machine learning: A review.
\newblock {\em ACM Computing Surveys}, 56(12):1--42, 2024.

\bibitem{artelt2019computation}
Andr{\'e} Artelt and Barbara Hammer.
\newblock On the computation of counterfactual explanations--a survey.
\newblock {\em arXiv preprint arXiv:1911.07749}, 2019.

\bibitem{guidotti2022counterfactual}
Riccardo Guidotti.
\newblock Counterfactual explanations and how to find them: literature review and benchmarking.
\newblock {\em Data Mining and Knowledge Discovery}, pages 1--55, 2022.

\bibitem{laber2023shallow}
Eduardo Laber, Lucas Murtinho, and Felipe Oliveira.
\newblock Shallow decision trees for explainable k-means clustering.
\newblock {\em Pattern Recognition}, 137:109239, 2023.

\bibitem{tavallali2021k}
Pooya Tavallali, Peyman Tavallali, and Mukesh Singhal.
\newblock K-means tree: an optimal clustering tree for unsupervised learning.
\newblock {\em The journal of supercomputing}, 77(5):5239--5266, 2021.

\bibitem{bertsimas2021interpretable}
Dimitris Bertsimas, Agni Orfanoudaki, and Holly Wiberg.
\newblock Interpretable clustering: an optimization approach.
\newblock {\em Machine Learning}, 110(1):89--138, 2021.

\bibitem{chasani2024unsupervised}
Paraskevi Chasani and Aristidis Likas.
\newblock Unsupervised decision trees for axis unimodal clustering.
\newblock {\em Information}, 15(11):704, 2024.

\bibitem{kaufman2009finding}
Leonard Kaufman and Peter~J Rousseeuw.
\newblock {\em Finding groups in data: an introduction to cluster analysis}.
\newblock John Wiley \& Sons, 2009.

\bibitem{kelly2024uci}
Markelle Kelly, Rachel Longjohn, and Kolby Nottingham.
\newblock The uci machine learning repository.
\newblock \url{https://archive.ics.uci.edu}.

\bibitem{mothilal2020explaining}
Ramaravind~K Mothilal, Amit Sharma, and Chenhao Tan.
\newblock Explaining machine learning classifiers through diverse counterfactual explanations.
\newblock In {\em Proceedings of the 2020 conference on fairness, accountability, and transparency}, pages 607--617, 2020.

\bibitem{van2021interpretable}
Arnaud Van~Looveren and Janis Klaise.
\newblock Interpretable counterfactual explanations guided by prototypes.
\newblock In {\em Joint European Conference on Machine Learning and Knowledge Discovery in Databases}, pages 650--665. Springer, 2021.

\end{thebibliography}

\appendix
\appendixpage

\section{DiCE and GuidedPrototypes}
In the following, we present the optimizations functions and parameters of {\dice} and {\prt}.
\subsection{DiCE parameters}
In order to generate a set $C=\{c_1, \ldots, c_k\}$ of counterfactuals, {\dice} minimizes the following objective function assuming a classification model $f$, a factual $x$ and target class $y$: 
\begin{align*}
C(x) & =  \arg \min_{c_1, \dots, c_k} 
    \frac{1}{k} \sum_{i=1}^{k} \text{yloss}(f(c_i), y) \\
    & + \frac{\lambda_1}{k} \sum_{i=1}^{k} \text{dist}(c_i, x) 
    - \lambda_2 \cdot \text{dpp\_diversity}(c_1, \dots, c_k)
\end{align*}
where $yloss(f(c_i),y)$ is the classification error for $c_i$, $dist(c_i,x)$ is the proximity between $c_i$ and $x$ and $dpp\_{diversity}$
promotes variety of the generated counterfactuals.
In our experiments, we use the following values determined after fine-tuning so as to generate the maximum possible number of counterfactual in the target class:
$total\_cfs=1$ (a single counterfactual is required, $k=1$), $\lambda_1=1$ (called $proximity\_weight$), $\lambda_2=0$ (called $diversity\_weight)$ and 
$stopping\_threshold=0.5$, indicating the minimum target class probability. 

\subsection{GuidedPrototypes parameters}
{\prt} mimizes the following objective function: \[
L = c \cdot L_{\text{pred}} + \beta \cdot L_1 + L_2 + \gamma L_{\text{AE}} + \theta L_{\text{proto}}
\] 
where the prediction loss $L_{pred}$ enforces the counterfactual to belong to the target class, while $L_1$ and $L_2$ denote distance norms between factual and counterfactual. The loss term $L_{AE}$ is used to keep the counterfactual in the data manifold by minimizing an autoencoder reconstruction error, while $L_{proto}$ guides counterfactuals towards the prototype of the target class for distribution alignment. Note that in {\prt} the dataset should be available.
The following parameters and corresponding values of the Alibi Prototypes framework were used:  
$\beta=0$, $\gamma=0$, $max\_iterations=5000$, the $kdtree$ option is selected and the
$feature\_range$ option is used to account for immutable features. Parameter $c\_init=1.0$ provides an initialization of $c$, and parameter $c\_steps$ defines the number of steps required to adjust $c$ during search. Parameters $\theta$ and $c\_steps$ required manual tuning to be properly specified for each experiment for attaining valid counterfactuals. The selected values are shown in Table \ref{tab:parameters-lr-qda}.

\begin{table}[htbp]
\centering
\scriptsize
\caption{{\prt} parameter values used in the experiments.}
\label{tab:parameters-lr-qda}
\begin{tabular}{lcccc} 
\toprule
Parameters & c\_steps(LR) & $\theta$(LR) & c\_steps(QDA) & $\theta$(QDA)   \\ 
\midrule
2D      &3 & 1.5 & 3 & 1.5 \\ 
2D Imm. Feat $1$ & 10 & 1.5 & 10 & 1.5\\
2D Imm. Feat. $2$ & 5 & 1.5 & 5 & 1.5\\
3D & 3 &  50 & 3 &  1.5\\
3D Imm. Feat. $2$ &10 & 1.5 & 7 & 1.5 \\
3D Imm. Feat $1,3$ &10  & 1.5 & 7  & 50\\
Iris & 3  &  1.5 & 3  &  50\\
Iris Imm. Feat. $1$ & 3 & 50 & 3 & 50\\
Iris Imm. Feat. $1,2$ & 3 &50 & 3 & 50\\
Wine & 3 & 1.5 & 3 & 1.5\\
\makecell[l]{Wine Imm. Feat. \\  $1,2,5,10$} & 3 & 1.5 & 3 & 1.5\\
\makecell[l]{Wine Imm. Feat. \\  $1,2,5,10-13$} & 3 & 1.5 & 3 & 50 \\

Pendigits & 3 & 1.5 &3 & 1.5 \\
\makecell[l]{Pendigits Imm. Feat. \\  $0,1,2$} & 3 & 1.5  & 5 & 1.5\\
\makecell[l]{Pendigits Imm. Feat. \\  $0,1,2,3,4,5$} & 3 & 1.5 & 5 & 1.5 \\
\bottomrule
\end{tabular}
\end{table}

\subsection{Counterfactuls in the target class}
In Table \ref{tab:success-lr-qda} , we present the percentage of generated solutions that belong to the target class for each case.

\begin{table}[htbp]
\scriptsize
\centering
\caption{Percentage of successfully generated counterfactuals (belonging to the target cluster).}
\label{tab:success-lr-qda}
\begin{tabular}{lcccc} 
\toprule
Method & PRT(LR) & DICE(LR) & PRT(QDA) & DICE(QDA) \\ 
\midrule
2D      & 100\% & 98\% & 100\% & 98\%  \\ 
2D Imm. Feat. $1$ & 84\% & 92\% & 100\% & 98\%  \\
2D Imm. Feat. $2$ & 60\% & 98\% & 100\% & 98\% \\
3D &  82\% & 100\% & 100\% & 100\% \\
3D Imm. Feat. $2$ & 100\% & 76\% & 100\% & 100\% \\
3D Imm. Feat. $1,3$ &60\% & 90\% & 100\% &64\% \\
Iris &100\% & 94\%& 100\%& 100\% \\
Iris Imm. Feat. $1$ & 30\% & 100\%& 100\% & 100\%\\
Iris Imm. Feat. $1,2$ & 20\% & 100\%&  100\% & 100\% \\
Wine & 100\% & 100\% & 100\% &  100\%\\
\makecell[l]{Wine Imm. Feat. \\  $1,2,5,10$}  & 100\% & 100\%& 100\% &  100\% \\
\makecell[l]{Wine Imm. Feat. \\  $1,2,5,10-13$} & 100\% & 100\% &100\% & 0\% \\
Pendigits & 100\% & 100\% & 96\% & 100\% \\
\makecell[l]{Pendigits Imm. Feat. \\  $0,1,2$} & 100\% & 100\% & 92\% &  100\% \\
\makecell[l]{Pendigits Imm. Feat. \\  $0,1,2,3,4,5$} & 100\% &100\% & 76\% & 100\%  \\
\bottomrule
\end{tabular}
\end{table}

\end{document}